\documentclass[journal]{IEEEtran}
\usepackage[ruled]{algorithm2e}
\usepackage{algorithmic}
\usepackage{color,xcolor,soul}
\usepackage{graphicx}
\usepackage{epstopdf}
\usepackage{mhchem}
\usepackage{subfigure}
\usepackage{bm}
\usepackage{tabularx}
\usepackage{colortbl,booktabs}
\usepackage{multirow}
\usepackage{adjustbox}
\usepackage{array} 
\usepackage{makecell}
\usepackage[colorlinks,linkcolor=blue]{hyperref}

\newcommand{\PreserveBackslash}[1]{\let\temp=\\#1\let\\=\temp}
\newcolumntype{C}[1]{>{\PreserveBackslash\centering}p{#1}}
\newcolumntype{R}[1]{>{\PreserveBackslash\raggedleft}p{#1}}
\newcolumntype{L}[1]{>{\PreserveBackslash\raggedright}p{#1}}

\begin{document}
	\title{GBSVM: Granular-ball Support Vector Machine}

	\author{Shuyin~Xia, Xiaoyu~Lian,
		Guoyin~Wang*,
		Xinbo Gao,
		Jiancu~Chen*,
		Xiaoli Peng

		\IEEEcompsocitemizethanks{\IEEEcompsocthanksitem S. Xia, X. Lian, G. Wang, J. Chen, X. Pemg \& X. Gao are with the Chongqing Key Laboratory of Computational Intelligence, Key Laboratory of Big Data Intelligent Computing, Key Laboratory of Cyberspace Big Data Intelligent Security, Ministry of Education, Chongqing University of Posts and  Telecommunications, 400065, Chongqing, China. E-mail: xiasy@cqupt.edu.cn, 1258852995@qq.com, wanggy@cqupt.edu.cn, gaoxb@cqupt.edu.cn, chenchen2153@163.com, 93334586@cqupt.edu.cn. \protect\\
		}
	}
	
	\markboth{ }%
	{Shell \MakeLowercase{\textit{et al.}}: Bare Demo of IEEEtran.cls for Computer Society Journals}
	
	
	\maketitle
	
	\begin{abstract}
		GBSVM (Granular-ball Support Vector Machine) is a significant attempt to construct a classifier using the coarse-to-fine granularity of a granular-ball as input, rather than a single data point. It is the first classifier whose input contains no points. However, the existing model has some errors, and its dual model has not been derived. As a result, the current algorithm cannot be implemented or applied. To address these problems, this paper has fixed the errors of the original model of the existing GBSVM, and derived its dual model. Furthermore, a particle swarm optimization algorithm is designed to solve the dual model. The sequential minimal optimization algorithm is also carefully designed to solve the dual model. The solution is faster and more stable than the particle swarm optimization based version. The experimental results on the UCI benchmark datasets demonstrate that GBSVM has good robustness and efficiency. All codes have been released in the open source library at http://www.cquptshuyinxia.com/GBSVM.html or https://github.com/syxiaa/GBSVM.
	\end{abstract}
	

	\IEEEpeerreviewmaketitle

	\vspace{-1em}	
	\section{Introduction}
	\IEEEPARstart{H}uman cognition has the characteristic of global precedence, i.e., from large to small, coarse to fine \cite{1982Topological}. Human beings have the ability of granulating data and knowledge into different granularity according to different tasks, and then solve problems using the relation between these granularity. Since Lin and Zadeh proposed granular computing in 1996, more and more scholars began to study information granular \cite{2018Multi,2021miaocla,2014Witold}, which simulates human cognition to deal with complexity problems \cite{2019Miao,2021Miaoclassi}. Granular computing advocates observing and analyzing the same problem from different granularity. The coarser the granularity, the more efficient the learning process and the more robust to noise; while the finer the granularity, the more details of things can be reflected. Choosing different granularity according to different application scenarios can more effectively solve practical problems \cite{2015miao,xiaGB,2022solve,2021Trend}. Wang introduced the cognitive law of ``global precedence'' into granular computing and proposed multi-granularity cognitive computing \cite{1982Topological,2018A}. 
	
\begin{figure}[ht]
		\centering
		\includegraphics[width = 0.2\textwidth]{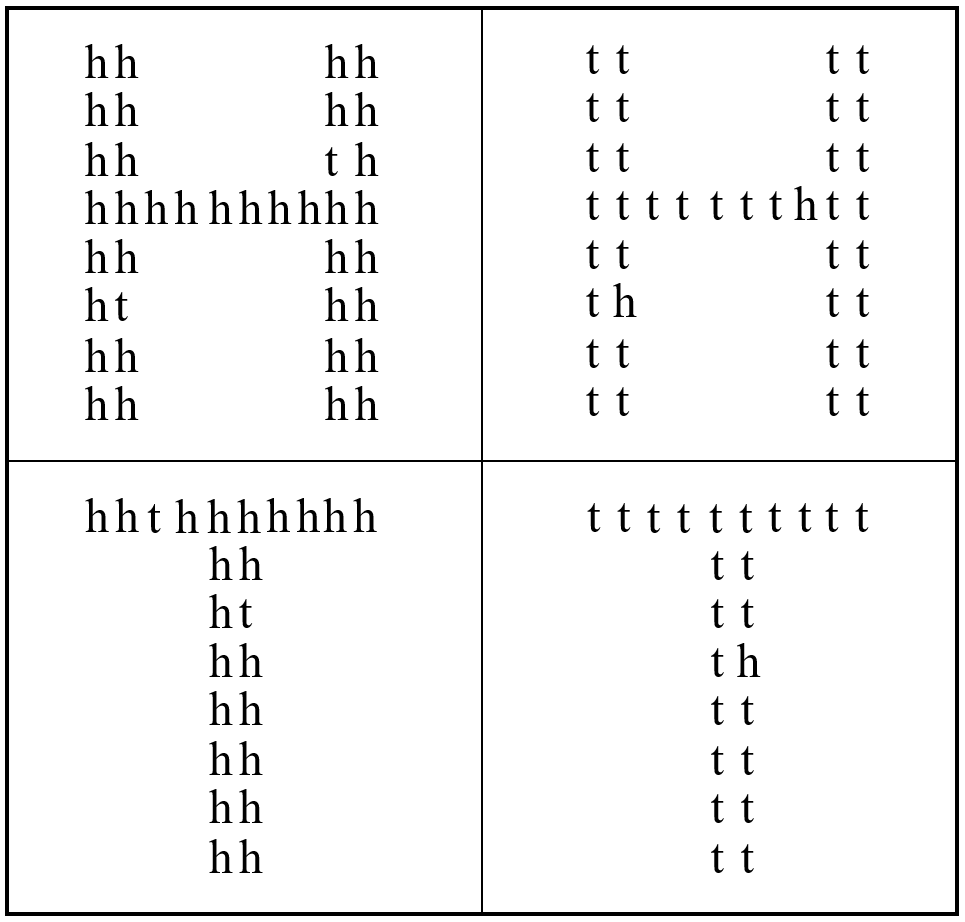}\\
		\caption{Human cognition ``global precedence".}
		\label{fig:cognition}
	\end{figure}

	Based on multi-granularity cognitive computing, Xia and Wang further proposed granular-ball computing by partitioning the dataset into some hyper-balls with different sizes (i.e., different granularity), called granular-balls \cite{xiaGB}. The reason why a hyper-ball is used instead of other geometries is that it has completely symmetrical geometric characteristics and a simple representation, i.e., that it only contains two parameters including the center and radius in any dimension. So, it is suitable for scaling up to high dimensional data. Other geometries do not have this characteristic. To simulate the cognitive law of ``global precedence'', granular-balls are generated by splitting the initial granular-ball of a whole data set from coarse to fine. As granular-balls adaptively have different sizes, they can fit various datasets with complex distribution. Instead of the finest granularity of a data point, a granular-ball can describe the coarse features of the covered samples. Different from the traditional AI learning methods whose input representation is the finest granularity of a point input, based on the granular-ball representation, granular-ball computing needs to create new computation models for various AI fields, such as classification, clustering, optimization, artificial neural networks and others. As the number of coarse granular-balls is much smaller than the finest data points, granular-ball computing is much more efficient than traditional AI computations; in addition, as a granular-ball is coarse and not easily affected by noise points, it is robust; furthermore, in comparison with a data point, a granular-ball can represent a point set and describe an equivalence class, it has better interpretability. In summary, Xia and Wang find that the cognitive law of ``global precedence'' and granular-ball computing have good computation performance in efficiency, robustness and interpretability at the same time \cite{xiaGB,xia2023granular}. These characteristics can be described in Fig. \ref{fig:cognition}, in the cognitive law of ``global precedence'', the human brain does not need to see the details or information of each point when recognizing large ``H'' and large ``T''; however, existing convolutional neural networks need to first convert the image into a pixel matrix, the finest granularity, and then calculate the contour information of large ``H'' and large ``T'' based on the pixel matrix. Obviously, the former is efficient. In addition, ``H'' and ``T'' are seen first, and then the ``h'' and ``t'' that make up them. There are two ``t''s in the left ``H'', which can be considered as noise, but it still does not affect the overall appearance of ``H''. Therefore, the cognitive law is robust. Finally, human recognition is based on semantic ``point sets'' such as ``lines'', rather than the finest granularity of ``points'' without any semantics. Therefore, the recognition process is interpretable. As a ``global precedence'' cognitive method, the characteristics of granular-ball computing and the aforementioned cognition are completely consistent.

	\begin{figure}[!htbp]
		\centering
		\subfigure[The existing classifiers]{\includegraphics[width =0.228\textwidth]{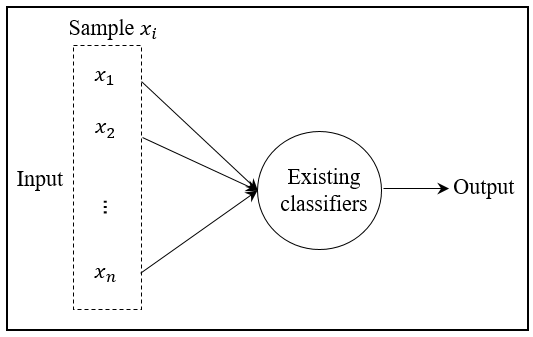}}
		\subfigure[The granular-ball classifiers]{\includegraphics[width =0.228\textwidth]{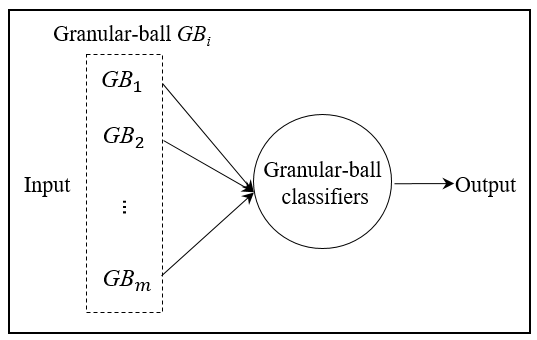}}
		\caption{The comparison of the granular-ball classifiers and the existing classifiers.}
		\label{fig:input VS}
	\end{figure}
	
		The origin of granular-ball computing was originally used to solve the problem of traditional classifiers not being able to achieve multi granularity learning. The existing classifiers such as traditional SVM process data with the finest granularity, i.e., a sample point or pixel point, as in Fig. \ref{fig:input VS}(a). Consequently, those classifiers have the disadvantages of low efficiency and low anti-noise ability \cite{2020miaoML,2022XiaPami}. To address this problem, Xia et al. proposed the granular-ball support vector machine (GBSVM) using the granular-balls generated on the dataset as its input instead of the data points \cite{xiaGB}, as shown in Fig. \ref{fig:input VS}(b). As granular-ball computing has the advantages of robustness, high efficiency and interpretability, the GBSVM exhibits good performance in both efficiency and robustness. Although existing GBSVM creates a non point input model, the existing version has three disadvantages: firstly, the existing GBSVM model has some minor errors; secondly, the dual model of GBSVM and nonlinear GBSVM model are not derived; thirdly, how to solute GBSVM is not provided. The work in \cite{xue2021dual} is also developed based on the wrong GBSVM, and there are some errors in its derived dual model. The contributions of this paper are as follows:
	\begin{enumerate}[\IEEEsetlabelwidth{4)}]
		\item The existing GBSVM model has an error that the support vector is not contained in the constraint. We have corrected it in the proposed model in this paper.
		\item The dual model of GBSVM is derived. It has a completely consistent formulation with the SVM.
		\item The granular-ball generation method in kernel space is developed, so the nonlinear GBSVM model is designed.
		\item A solution algorithm for the dual model using the particle swarm optimization (PSO) algorithm \cite{flake2002efficient} and sequential minimal optimization (SMO) algorithm \cite{platt1998sequential} are designed. The experimental results on some benchmark data sets demonstrate better performance of GBSVM in processing classification problems with label noise in comparison with SVM.
	\end{enumerate}
	
	The rest of this paper is organized as follows. We introduce related works in Section \ref{sec:relatedwork}. The model and dual model of GBSVM are presented in detail in Section \ref{sec:GBSVM}. Experimental results and analysis are presented in Section \ref{sec:experiment}. We present our conclusion and future work in Section \ref{sec:conclusion}.
	
	\section{Related Work}{\label{sec:relatedwork}}
	\subsection{Granular-ball Computing}{\label{sec:GBC}}
	In the original works about classification and clustering based on granular-ball computing, a granular-ball $GB$ is defined as the following standard form: ${GB}_j=\{x_{i}|i=1,2,\dots,k\}$, and its center $c=\frac{1}{k}\sum_{i=1}^{k}x_i$, where $x_i$ and $k$ denote a sample in the granular-ball and the number of samples in the granular-ball, respectively. The radius can be obtained in many ways. Two main ways are the average distance and maximum distance. The average distance is $r=\frac{1}{k}\sum_{i=1}^{k}||x_i-c||$, i.e., the average distance between $c$ to the other samples in a granular-ball. The maximum distance is $r=\max||x_i-c||$, i.e., the maximum distance between $c$ to the other samples in the granular-ball. The granular-balls generated with the average distance can fit the data distribution well and get a clearer decision boundary than that with the maximum distance. The granular-balls generated with the maximum distance can cover all samples in the sample space than that with the average distance. The above definition of a granular-ball apply to classification and may differ in other fields, such as optimization and graph representation \cite{xia2023Optimization, shuyin2023graph}. The granular-ball computing can be described as the following model.

	\begin{figure}[!htbp]
	\centering
	\includegraphics[width = 0.418 \textwidth]{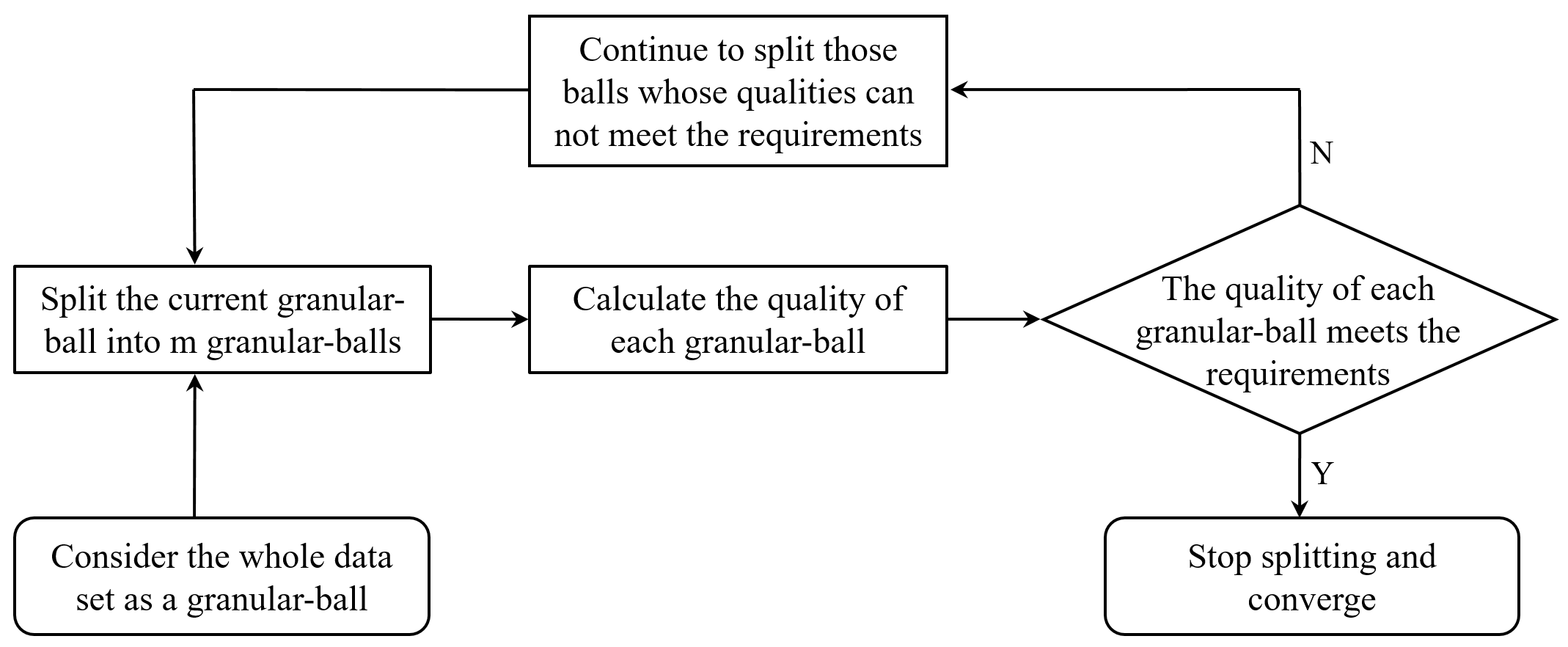}\\
	\caption{Process of the granular-ball generation.}
	\label{fig:GBgeneration}
\end{figure}

	\textcolor{black}{Given a dataset $D$ with $n$ samples, $x_i \in D$.} $GB_j(j=1,2,\dots,\textcolor{black}{m})$ is the granular-ball generated on \textcolor{black}{the dataset $D$}, where $m$ denotes the number of granular-balls \cite{2021GBAdai}. The optimization model is formulated as:
	\begin{equation}\label{equ:GBC}
		\begin{aligned}
			&f(x_i,\vec{\alpha}) \longrightarrow  g(GB_j,\vec{\beta}),\\
		s.t. \quad	&\min \quad \frac{\textcolor{black}{n}}{\sum_{j=1}^{\textcolor{black}{m}}{|GB_j|}}+ \textcolor{black}{m}+loss(GB_j),\\
			&quality(GB_j)\geq T,
		\end{aligned}
	\end{equation}
	where $|\cdot|$ is the cardinality, i.e., the number of objects \textcolor{black}{contained} in a set. The function $f$ means the original traditional learning methods that take points $x$ as input, while function $g$ stands for the granular-ball learning methods that utilize granular-balls $GB$ as input. Overall, granular-ball computing encompasses two aspects: the novel multi granularity representations using granular-ball, and the novel computing modes based on granular-ball representation. In the objective of the constraints, the first term indicates the sample coverage; the higher the coverage, the better, and the less information loss. The second term $m$ represents the number of granular-balls. The fewer the number of granular-balls, the better under the constraint: the more efficient and robust the calculation of granular-balls. The third term $loss(GB_j)$ is used to optimize the granular-ball quality in the cases where the point representation is changed in the training process, such as in deep neural networks; otherwise, in traditional learning tasks, such as GBSVM, this term is not needed. In the constraints of the constraints of (\ref{equ:GBC}), the quality of each granular-ball, i.e. purity, should meet a threshold $T$. Only the parameter $T$ is introduced in the constraints in the constraints of Model (\ref{equ:GBC}), and the parameter can further be eliminated using also adaptive rules \cite{xia2022efficient}; in many other granular-ball methods, such as granular-ball clustering, the parameter also does not \textcolor{black}{exist} \cite{xia2023granular}. So, the optimization algorithm for granular-ball generation is very \textcolor{black}{adaptable}. The granular-ball becomes finer when the radius $r$ decreases and coarser when $r$ increases. In classification problems, not all data points have to belong to a particular granular-ball with guaranteed quality of each ball, which is beneficial for generating a clearer decision boundary in some cases; in most other problems, such as granular-ball clustering, the radius of a granular-ball is equal to the maximum distance from the points inside it to its center so that granular-balls can cover all data points \cite{xia2023granular}. In supervised tasks, such as classification, purity is used as one of the evaluation standards to measure the quality of the granular-ball. The purity is equal to the proportion of the majority of samples in a granular-ball \cite{xiaGB}.

\begin{figure}[!htbp]
		\centering
		\subfigure[]		{\includegraphics[width =1.6in]{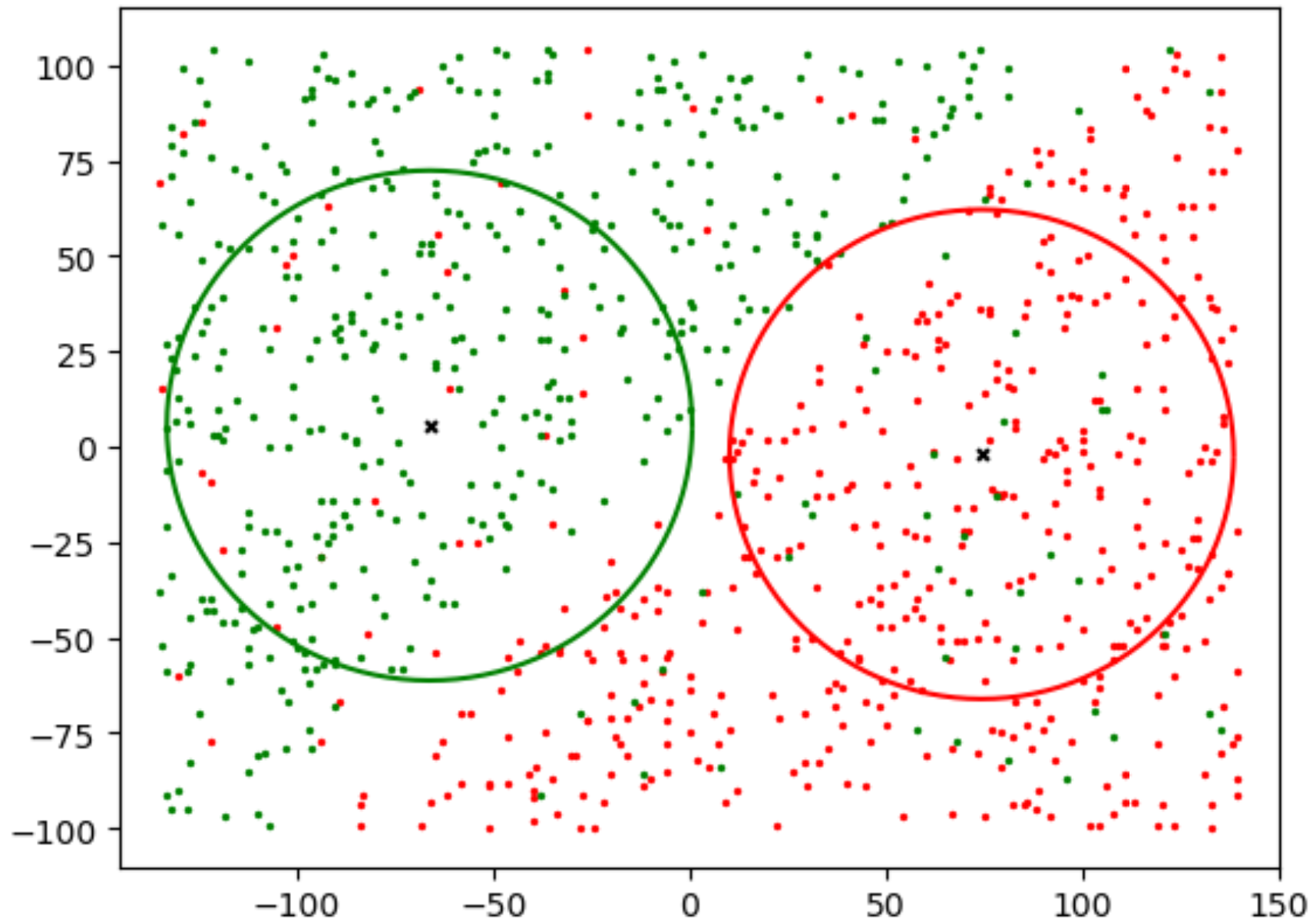}}
		\subfigure[]		{\includegraphics[width =1.6in]{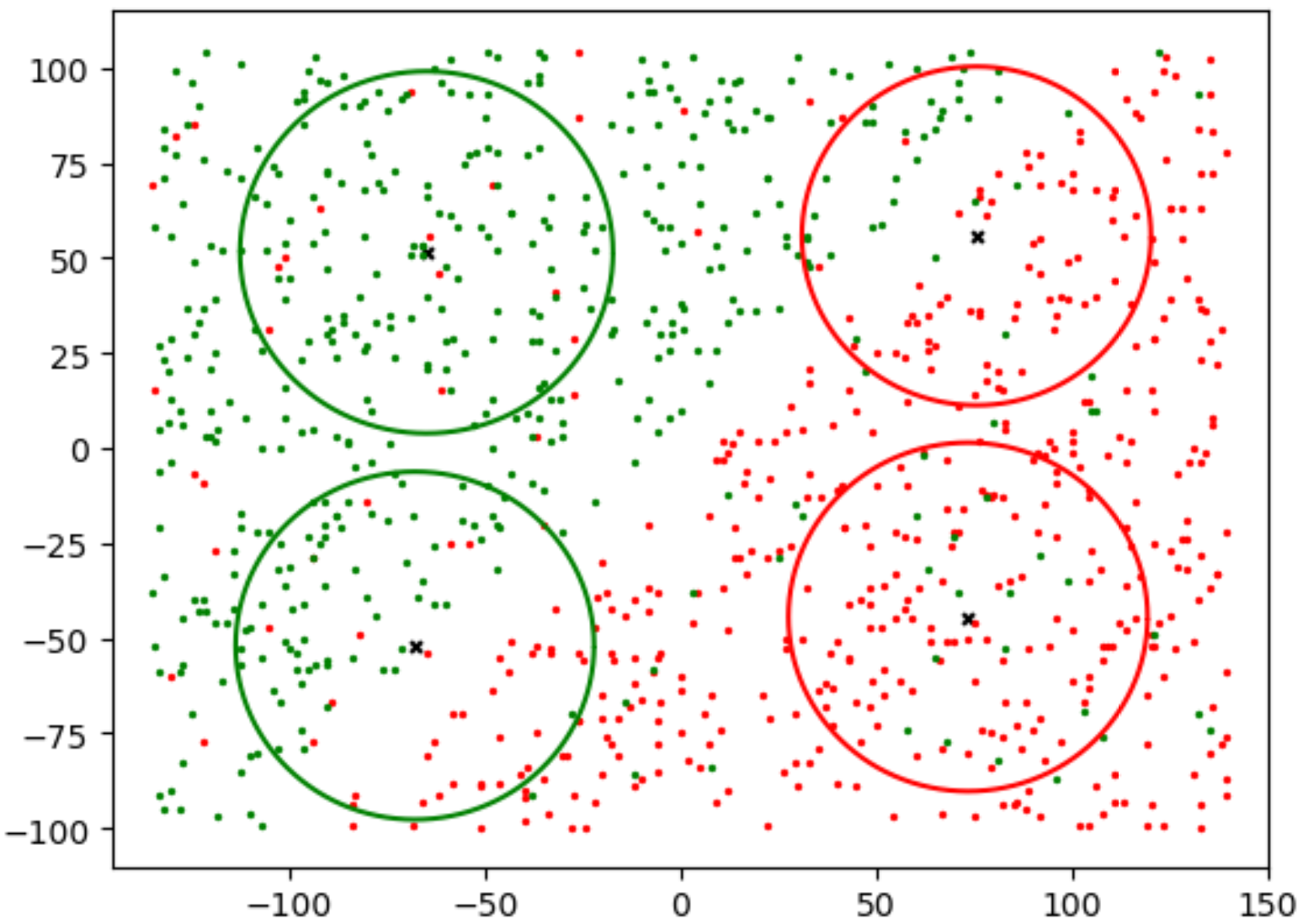}}
		\subfigure[]		{\includegraphics[width =1.6in]{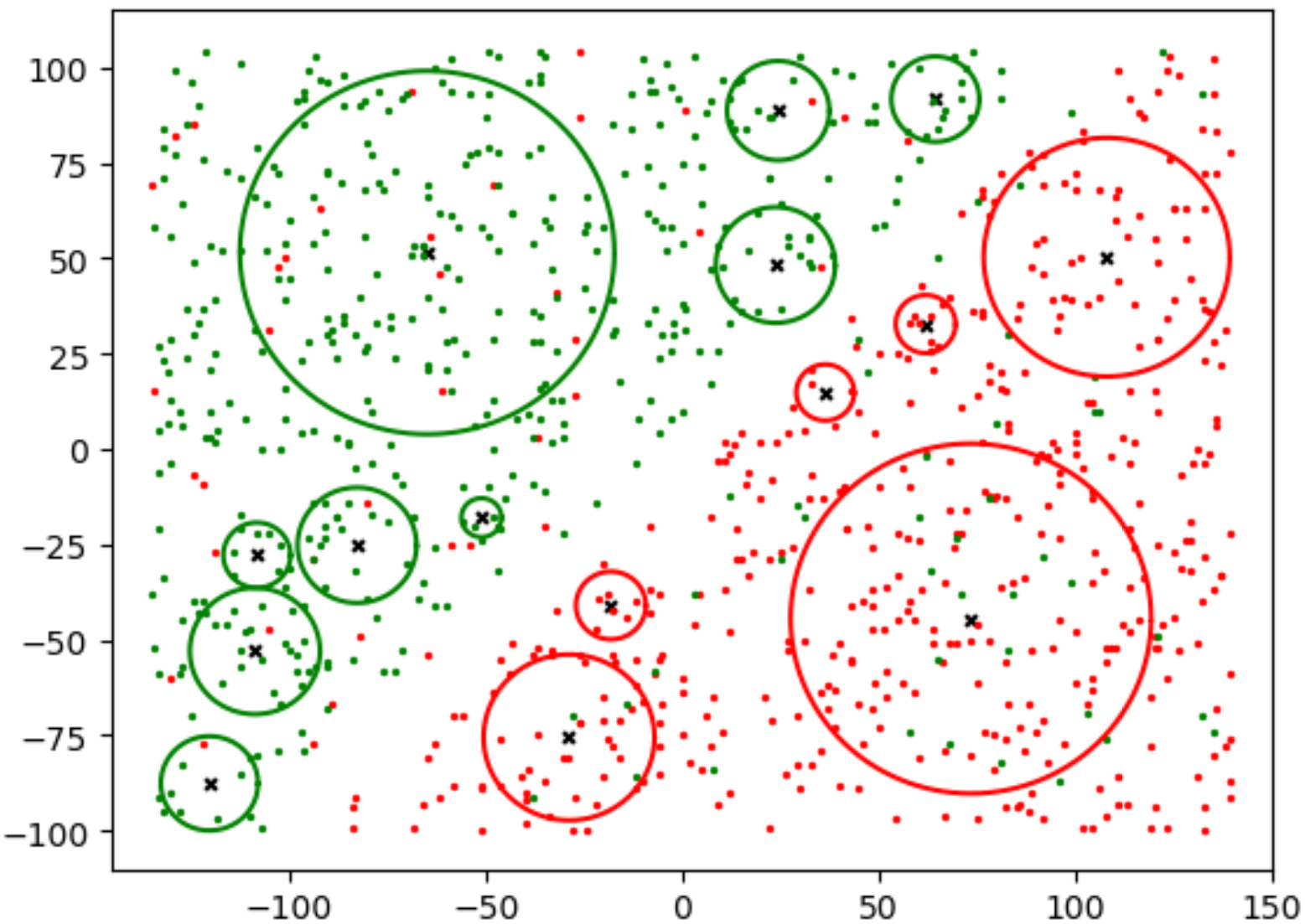}}
		\subfigure[]		{\includegraphics[width =1.6in]{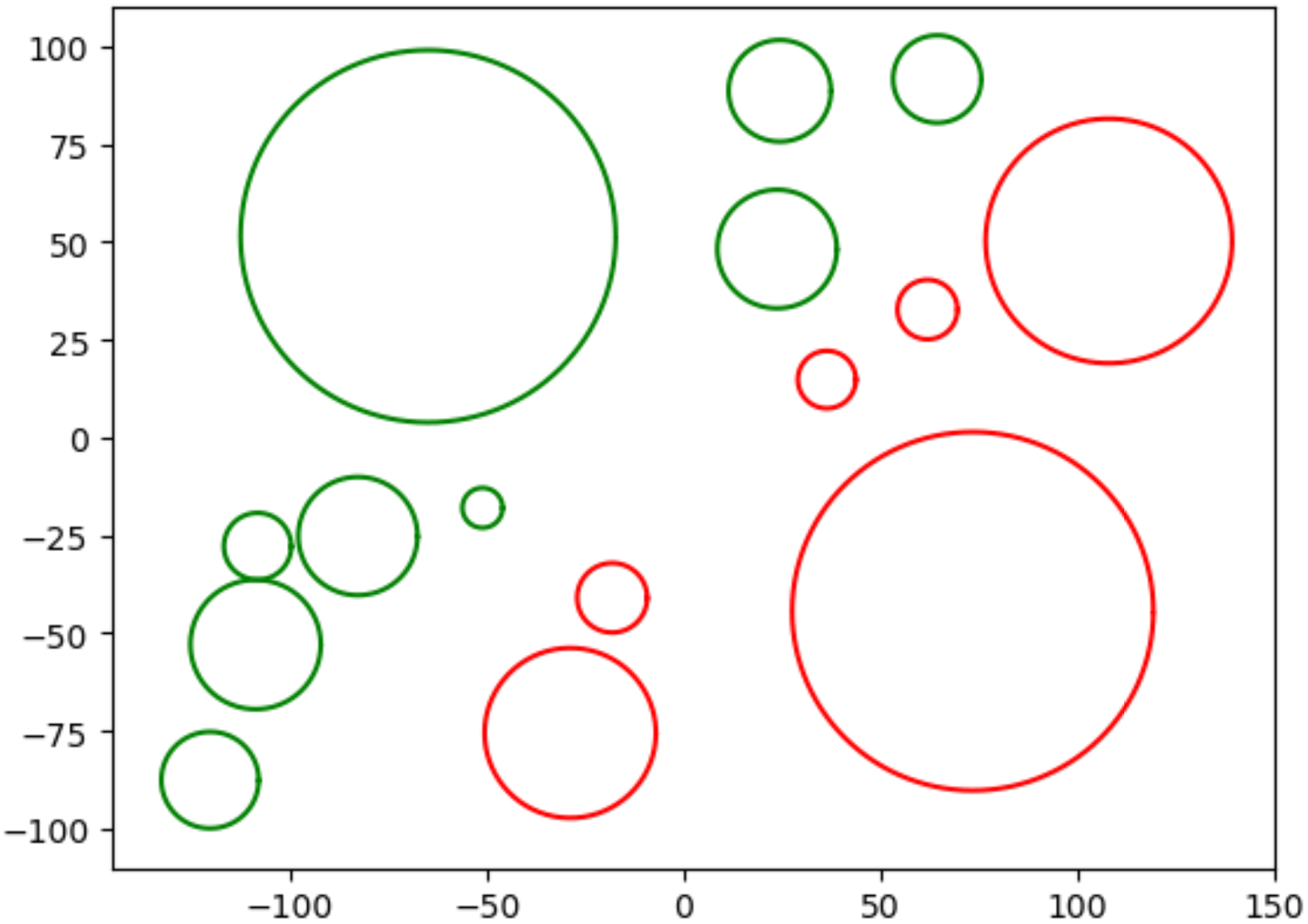}}
		\caption{The splitting process of the granular-ball. (The purity threshold is 0.9, samples and granular-balls of two color represent two classes.) (a) The granular-balls generated in the 2$^{nd}$ iteration. (b) The granular-balls generated in the 3$^{rd}$ iteration. (c) The stable granular-balls. (d) The final granular-balls extracted from (c).}
		\label{fig:splitting}
	\end{figure}

Fig. \ref{fig:GBgeneration} shows a heuristic solution strategy for (\ref{equ:GBC}). Taking classification as an example, the process of the granular-ball generation is shown in Fig. \ref{fig:GBgeneration}. To simulate the ``global precedence'' of human cognition, the whole dataset is regarded as an initial granular-ball, whose purity is the worst and can not describe the distribution of the dataset. When the purity of a granular-ball is lower than the purity threshold, the quality of this granular-ball does not meet the condition and needs to continue being split from coarse to fine until the purities of its sub-granular-balls are no less than the purity threshold. The higher the purity of the generated granular-balls, the better the consistency with the data distribution of the original dataset. Each granular-ball also has a label, which is consistent with the label of the majority \textcolor{black}{of} samples in the granular-ball. The process of the granular-ball generation is as Fig. \ref{fig:GBgeneration}. As shown in Fig. \ref{fig:splitting}(a), there are two classes, so the initial granular-ball is split into two granular-balls using k-means or k-division. As the purity of each granular-ball is not high enough, the two granular-balls continue to be split. As the splitting process continues, the purity of each granular-ball increases, and the decision boundary becomes clear until the purities of all granular-balls meet the stop condition as shown in Fig. \ref{fig:splitting}(c). Fig. \ref{fig:splitting}(d) shows the final granular-balls extracted from Fig. \ref{fig:splitting}(c). \textcolor{black}{It is worth noting that the radius of each ball in Fig. \ref{fig:splitting} is the average of all points in the ball, so all sample points are not fully covered.}
	
	In the granular-ball classifier, the learning process includes the granular generation and granular-ball computing learning. Since the number of granular-balls can be considered almost as a small constant, the time complexity of granular-ball generation is equal to the time complexity of generating the largest granular-ball among. 2-mean clustering or k-devision \cite{xia2022efficient} generally has a fast convergence rate, so it can be considered an approximately linear algorithm. Since the size of the data set containing the centers and radius of granular-balls is very small, the regression process can be almost negligible. Therefore, the time complexity of the granular-ball classifier is reduced to close to $O (N)$, which is very low. It should be noted that, the representation of granular-balls may not necessarily use strict balls in some scenarios, but can be approximated in some ways, such as using rectangles in the granular-ball image processing \cite{shuyin2023graph} and using the boundary points of balls in granular-ball evolutionary computation \cite{xia2023Optimization}.
  
	Although the theory of granular-ball computing has not been proposed for a long time, a series of methods have been developed to address problems such as efficiency, robustness, and interpretability in various fields of artificial intelligence. To address the low efficiency and hyper-parameter (i.e., the neighborhood radius) optimization problem in neighborhood rough set, The granular-ball neighborhood rough set (GBNRS) is proposed by introducing granular-ball computing into the neighborhood rough set \cite{GBNRS}. GBNRS is the first rough set method that is parameter-free in processing continuous data.  Furthermore, a unified model called granular-ball rough set (GBRS) is developed to address the conflicting problem in equivalent class knowledge representation and continuous data processing between Pawlak rough set and neighborhood rough set \cite{GBRS}. GBRS exhibits a higher accuracy than both the traditional neighborhood rough set and Pawlak rough set. Granular-ball computing is also introduced into sampling called granular-ball sampling (GBS), which is the first sampling method that can not only reduce the size of a data set but also improve the data quality in noisy label classification \cite{2021GBS}.  In short, granular-ball computing \cite{xia2023granular} has now been effectively extended to various fields of artificial intelligence, developing theoretical methods such as granular-ball clustering methods \cite{xie2023efficient}, granular-ball neural networks \cite{shuyin2023graph}, and granular-ball evolutionary computing \cite{xia2023Optimization}, significantly improving the efficiency, noise robustness, and interpretability of existing methods.
	
	\subsection{The Basic Model of GBSVM}
	SVM is a widely used classifier \cite{platt1998sequential}. There has been a lot of work on SVM to improve efficiency or robustness. Zhao et al. \cite{zhao2014efficient} proposed an efficient learning mechanism to build fuzzy rule-based systems through the construction of sparse least-squares support vector machines (LS-SVMs). By introducing some key techniques to avoid matrix inversion, the model speeds up the training process, which is an efficient algorithm. Twin support vector machine (TWSVM) \cite{khemchandani2007twin,ding2017twin} has gained considerable attention in the field of machine learning in recent years. The underlying concept of TWSVM involves seeking two non-parallel proximal hyperplanes that are closer to each of the two classes, thereby improving the generalization ability of generalized eigenvalue proximal SVM \cite{fung2001proximal,shao2012improved}. In particular, Chen et al. \cite{chen2019novel} proposed projective twin SVM (PTSVM), which identifies an optimal discriminant subspace for each class that minimizes the within-class scatter in the new subspace. Moreover, Chen et al. \cite{chen2020nu} further refined PTSVM into a novel non-parallel classifier called $v$-PTSVM. Tanveer et al. \cite{tanveer2022intuitionistic} proposed a novel intuitionistic fuzzy weighted least squares TWSVMs for cla ssification problems. The model uses local neighborhood information between data points, membership and non-membership weights to reduce the influence of noise and outliers. Ye et al. \cite{ye2021multiview} introduced a set of double-sided constraints to develop multiview robust double-sided twin SVM with SVM-type problems to promote classification performance. Furthermore, SVM has been applied to classification problems in various fields, and there have been several studies on this topic \cite{wang2018transfer,jia2020semisupervised,ren2022robust,wang2021twin}. These SVM methods can improve efficiency or robustness to an extent. However, most of them can't achieve good performance in both efficiency and robustness, one important reason is that the methods are all designed based on point input instead of multi-granularity representation; in addition, most of them are specific models instead of a general framework. Addressing these problems, the GBSVM model introduces granular-ball computing to generate granular-balls with different sizes as input, which greatly improves greatly enhances the performance of efficiency and robustness. Furthermore, this idea serves as a general framework that can be adapted to other algorithms.

\begin{figure}[!ht]
		\centering
		\subfigure[]		{\includegraphics[width =2.5in]{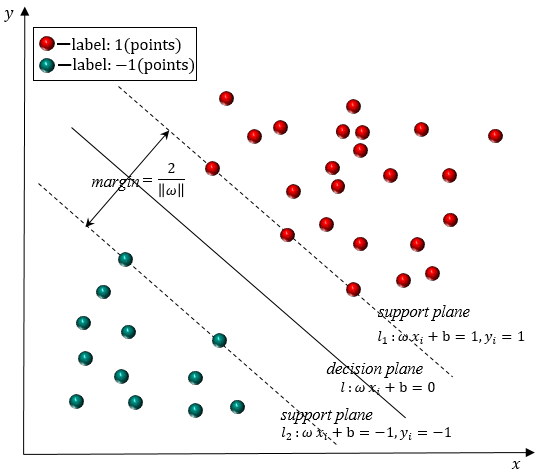}}
		\subfigure[]		{\includegraphics[width =2.5in]{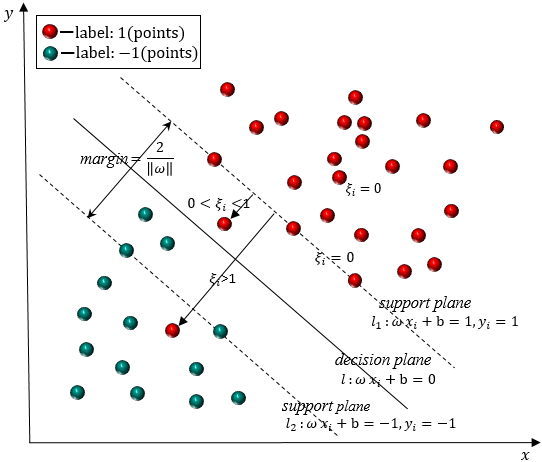}}
		\caption{Schematics of the traditional SVM. (a) The separable SVM; (b) The inseparable SVM.}
		\label{fig:SVM}
	\end{figure}

     The basic idea of traditional SVM is shown in Fig. \ref{fig:SVM}. Given a dataset $D=\{ (x_i,y_i), i=1,2,\dots, \textcolor{black}{n}, y_i \in \{+1,-1\} \}$, where $y_i$ denotes the label of $x_i$. In Fig. \ref{fig:SVM}, two colors represent two classes, i.e., ``+1'' and ``-1''. $l$ represents the optimal plane, also called the decision plane. $l_1$ and $l_2$ are the support planes parallel to $l$. The distance between $l_1$ and $l_2$ is the $margin$. $\omega$ and $b$ are the unit normal vector and the offset of $l$, respectively. In the linear separable case, the separable SVM is also called ``hard interval support vector machine'' as shown in Fig. \ref{fig:SVM}(a). Its learning strategy is to maximize the ``margin'' interval and formalize it into convex quadratic programming. The objective function of linear support vector machine in the separable classification is:   
     	\begin{equation}\label{equ:lSVM}
		\begin{aligned}
			\substack{\min\\\omega,b}\quad\quad &\frac{1}{2} \left\| \omega  \right\|^2,\\
			s.t.\quad\quad &y_i(\omega \cdot x_i +b)\ge 1 \text,\quad i=1,2,\cdots.
		\end{aligned}
	\end{equation}
	The dual model of Eq. (\ref{equ:lSVM}) is:
	\begin{equation} 
		\setlength{\abovedisplayskip}{3pt}
		\begin{aligned}
			\substack{\max\\\alpha}\quad\quad -&\frac{1}{2} \sum\limits_{i=1}^{\textcolor{black}{n}} \sum\limits_{j=1}^{\textcolor{black}{n}} \alpha_i \alpha_j y_i y_j x^{T}_{i} x_j+\sum\limits_{i=1}^{\textcolor{black}{n}}\alpha_i,\\
			s.t.\quad\quad &\sum\limits_{i=1}^{\textcolor{black}{n}}\alpha_iy_i=0 ,\\
			\quad\quad &\alpha_i\ge 0, \text,\quad i=1,2,\cdots.
		\end{aligned}
	\setlength{\belowdisplayskip}{3pt}
	\end{equation}

	The slack variable $\xi$ and penalty coefficient $C$ are imported into Eq. (\ref{equ:lSVM}) for addressing inseparable classification. Then, the inseparable support vector machine, i.e., ``soft interval support vector machine'' in Fig. \ref{fig:SVM}(b) is obtained. The objective function is transformed to be (\ref{equ:nonlSVM}).
	\begin{equation}\label{equ:nonlSVM}
		\begin{aligned}
			\substack{\min\\\omega,b}\quad\quad &\frac{1}{2} \left\| \omega  \right\|^2+C\sum\limits_{i=1}^{\textcolor{black}{n}}\xi_i,\\
			s.t.\quad\quad &y_i(\omega \cdot x_i +b)\ge 1-\xi_i \text,\quad i=1,2,\cdots.
		\end{aligned}
	\end{equation}

	The dual model of Eq. (\ref{equ:nonlSVM}) is:
	\begin{equation}\label{equ:lSVM-dual}
		\begin{aligned}
			\substack{\max\\\alpha}\quad\quad -&\frac{1}{2} \sum\limits_{i=1}^{\textcolor{black}{n}} \sum\limits_{j=1}^{\textcolor{black}{n}} \alpha_i \alpha_j y_i y_j x^{T}_{i} x_j+\sum\limits_{i=1}^{\textcolor{black}{n}}\alpha_i,\\
			s.t.\quad\quad &\sum\limits_{i=1}^{\textcolor{black}{n}}\alpha_iy_i=0 ,\\
			\quad\quad&0\le \alpha_i \le C \text,\quad i=1,2,\cdots.
		\end{aligned}
	\end{equation}
	
	\begin{figure}[!htbp]
		\centering
		\includegraphics[width = 0.38 \textwidth]{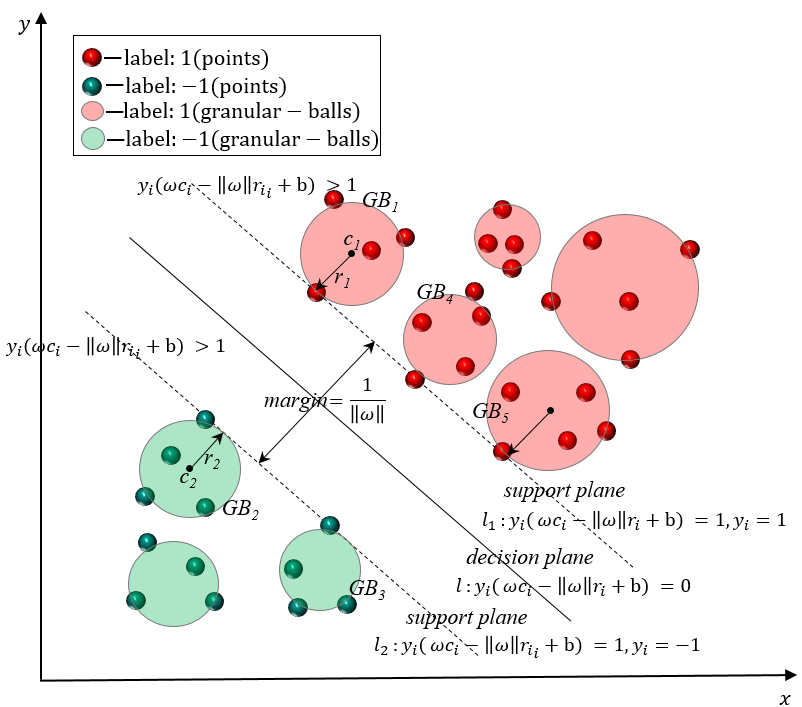}\\
		\caption{Schematic diagram of the old granular-ball support vector machine.}
		\label{fig:oldGBSVM}
	\end{figure}
	
	The schematic diagram of the existing GBSVM is presented in Fig. \ref{fig:oldGBSVM}. In GBSVM, data points are replaced with the granular-balls including the support granular-balls and non support granular-balls \cite{xiaGB}. The support vectors in the SVM model are replaced by the support granular-balls, e.g. $GB_1, GB_2, GB_3$ $GB_4$ and $GB_5$ in Fig. \ref{fig:oldGBSVM}; while the others by the non support granular-balls. The two support planes, i.e., $l_1$ and $l_2$, are constrained by two rules. The first one is that the support planes must be tangent to the support granular-balls; the second one is that the distance between the support plane and each granular-ball should be no less than the corresponding radius. A granular-ball containing some points is generated according to the method in Section \ref{sec:GBC}. Thus, the separable GBSVM model is:
	\begin{equation}\label{equ:oldGBSVM}
		\begin{aligned}
			\substack{\min\\\omega,b}\quad\quad &\frac{1}{2} \left\| \omega  \right\|^2\\
			s.t.\quad\quad &y_i(\omega \cdot c_i-||\omega||r_i+b) \geq 1 \text,\quad i=1,2,\cdots.
		\end{aligned} 
	\end{equation}
	In order to obtain the inseparable GBSVM model, the slack variables, i.e., $\xi_i$ and $\xi_i^\prime$, and the penalty coefficient $C$ are introduced. Then, the inseparable GBSVM model is derived as follows:
	\begin{equation}\label{equ:oldGBSVM-i}
		\begin{aligned}
			\substack{\min\\\omega,b}\quad\quad &\frac{1}{2} \left\| \omega  \right\|^2+C\sum_{i=1}^{\textcolor{black}{m}}\delta_i,\\
			s.t.\quad\quad &y_i(\omega \cdot c_i-||\omega||r_i+b) \geq 1 - \delta_i \text,\quad i=1,2,\cdots\\
			\quad\quad &\xi_i \ge 0,
		\end{aligned} 
	\end{equation}
	where $\delta_i=y_i\xi_i^\prime+||\omega||\xi_i$. 
	The existing GBSVM model \cite{xiaGB} employs granular-balls instead of points as input, significantly enhancing efficiency and robustness. Although there are some errors in the method, it is still introduced here.  
	
	\section{Granular-ball Support Vector Machine}{\label{sec:GBSVM}}
	\subsection{Motivation}
	The existing classifiers use the finest information granularity, i.e., the sample or pixel point, as the input, so the existing classifiers lack robustness to label noise. So, many extra methods are developed to detect label noise or make the existing classifiers robust to label noise \cite{XiaCRF,XiaRD,liu2015classification}. In contrast, GBSVM can be robust to label noise without using any other technology because the coarse granularity of a granular-ball can remove the affection of the finest granularity of a label noise point \cite{xiaGB}. However, the basic model of the existing GBSVM has some errors. To describe it clearly, the derivation process of GBSVM is introduced as follows. When considering a particular point to be positive and negative, the support planes, i.e., $l_1$ and $l_2$ shown in Fig. \ref{fig:SVM}, are denoted as follows:
	\begin{align}
		l_1:\quad\quad  \omega \cdot x_i+b=1, y_i=+1 ,\label{equ:oldl1} \quad\\
		l_2:\quad\quad \omega \cdot x_i+b=-1, y_i=-1.\label{equ:oldl2}
	\end{align}
	From Eqs. (\ref{equ:oldl1}) and (\ref{equ:oldl2}), we have:
	\begin{align}
		y_i(\omega \cdot x_i + b)=1,\label{equ:oldl} \\ 
		\Rightarrow \omega \cdot x_i =y_i-b.\label{equ:oldlchange}
	\end{align}
	As the vertical distance from a support center $c_i$ to the support plane should be equal to the corresponding support radius $r_i$, we have:
	\begin{equation}\label{equ:oldrule}
		\frac{\omega \cdot (c_i-x_i)}{\|\omega\|}=r_i.
	\end{equation}
	Eq. (\ref{equ:oldlchange}) is substituted into Eq. (\ref{equ:oldrule}), so,
	\begin{equation}
		y_i(\omega \cdot c_i -\|\omega\|r_i +b)=1.
	\end{equation}
	As the centers of granular-balls are located outside of the support planes, the constraint of GBSVM is expressed as:
	\begin{equation}\label{equ:oldstrain}
		y_i(\omega \cdot c_i -\|\omega\|r_i +b) \geq 1,
	\end{equation}
	which is also shown in Eq. (\ref{equ:oldGBSVM}), and the inseparable GBSVM is also derived as shown in Eq. (\ref{equ:oldGBSVM-i}). 
	
	However, Eq. (\ref{equ:oldrule}) can only obtain a plane that is tangent with a granular-ball, and the constraint of (\ref{equ:oldstrain}) ensures that the plane is not intersected with any granular-ball. In addition, when a plane tangent with a granular-ball needs to be calculated, the right side of Eq. (\ref{equ:oldrule}) should be $|r_i|$, but only the positive case is considered in (\ref{equ:oldrule}). Therefore, Eq. (\ref{equ:oldrule}) is not sufficient to find the support planes. Besides, the dual model and nonlinear GBSVM model are not derived, and no solution algorithm is designed. In this paper, we will address these problems.
	
	\subsection{Linear Granular-ball Support Vector Machine}
 
	\begin{figure}[ht]
	\centering
	\includegraphics[width = 0.5\textwidth]{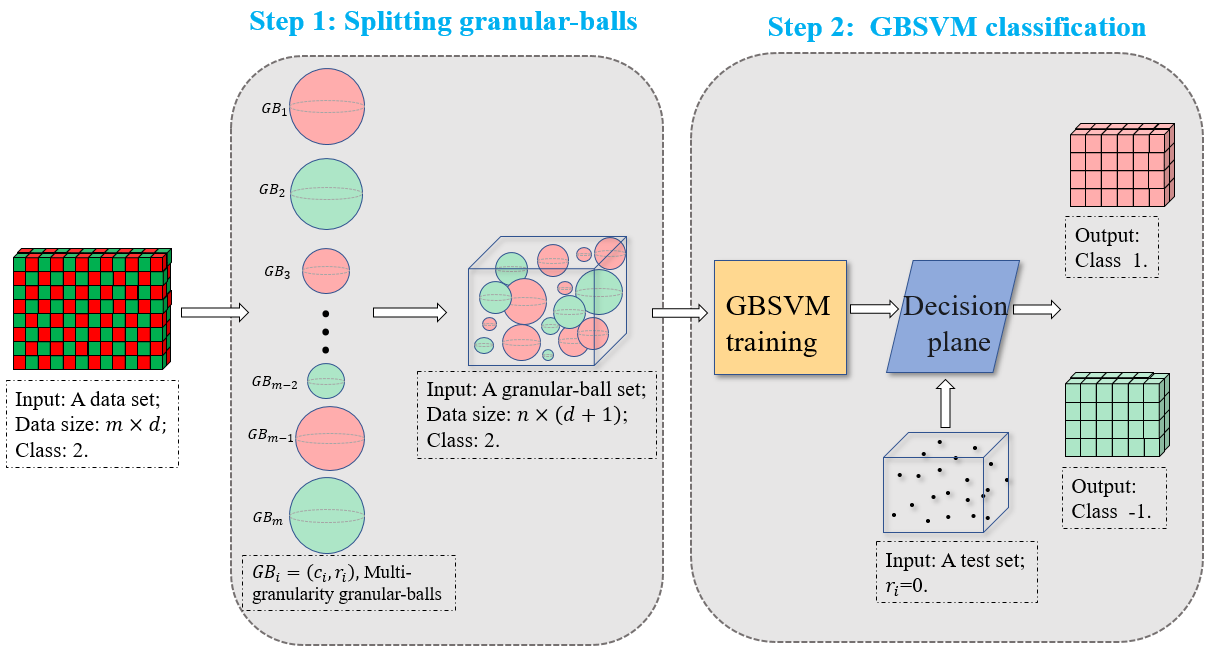}\\
	\caption{\textcolor{black}{Flow chart of GBSVM processing data.}} 
	\label{fig:Flow}
\end{figure}
	
\begin{figure}[!ht]
		\setlength{\abovecaptionskip}{-0.1cm}  
		\centering
		\subfigure[]{\includegraphics[width =3in]{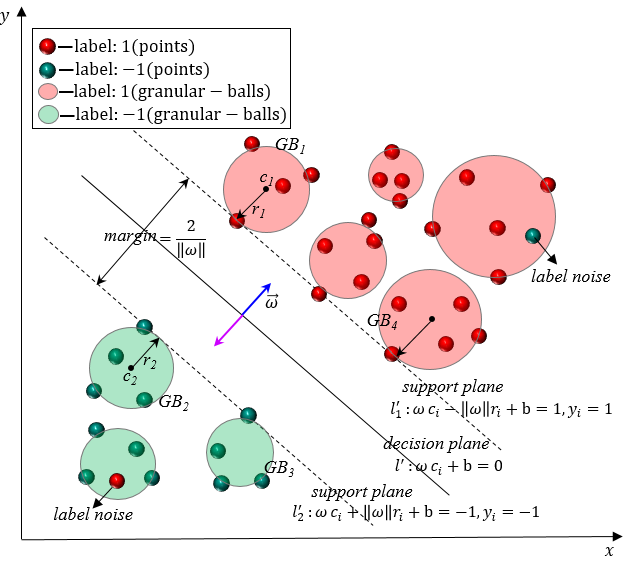}}
		\subfigure[]{\includegraphics[width =3in]{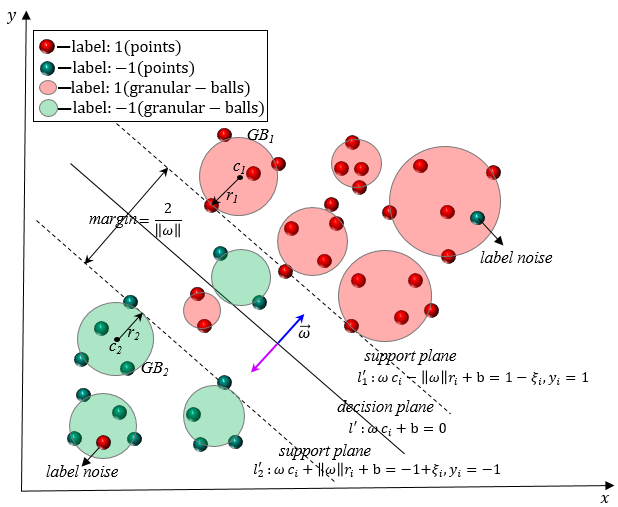}}
		\caption{Schematics of GBSVM. (a) The separable GBSVM; (b) The inseparable GBSVM.} 
		\label{fig:GBSVM}
	\end{figure}

	The GBSVM model initially splits the input data points into granular-balls of varying sizes. Subsequently, it feeds these balls, represented by their centers and radius, into the classifier to derive the classification model and outcomes, as illustrated in Fig. \ref{fig:Flow}. Notably, the classifier doesn't operate on individual points, $x_i$, but rather on the parameters, $c$ and $r$.
	Corresponding to Fig. \ref{fig:oldGBSVM}, the revised model of GBSVM is shown in Fig. \ref{fig:GBSVM}. 

	\subsubsection{The Separable GBSVM}
	\paragraph{The Original Model of the Separable GBSVM}
	As that in Fig. \ref{fig:SVM}(a), when considering a specific point as positive and negative, $l_1$ and $l_2$ are as follows:
	\begin{align}
		l_1:\quad\quad \omega \cdot x_i+b=1,y_i=+1,\quad \label{equ:l1} \\
		l_2:\quad\quad \omega \cdot x_i+b=-1,y_i=-1.\label{equ:l2}
	\end{align}
	Eq. (\ref{equ:l1}) and Eq. (\ref{equ:l2}) can be transformed to be:
	\begin{align}
		\omega \cdot x_i+b=y_i,\label{equ:l12} \\
		\Rightarrow \omega \cdot x_i=y_i-b.\label{equ:l12-n}
	\end{align}
	The schematic diagram of the separable GBSVM is shown in Fig. \ref{fig:GBSVM}(a). The direction of $\overrightarrow \omega$ is as the blue arrow and the purple arrow has an opposite direction. For the sake of simplicity, we use $\omega$ to represent $\overrightarrow \omega$. We set the direction of the blue arrow to the positive direction, and the purple to the negative. The heterogeneous points within a granular-ball represent label noise, which does not influence the overall label of the ball, demonstrating that granular-ball computing is robust. The red granular-balls are above $l_1^\prime$ and the same with the positive direction, so the distances from the red balls to $l_1^\prime$ are positive; while the green ones are below $l_2^\prime$ and opposite to the positive direction, the distances from these green balls to $l_2^\prime$ are negative. Suppose $l^\prime$ can correctly classify all granular-balls. The distance between the center of a granular-ball to the corresponding support plane must not be less than the radius of the corresponding support granular-balls, so the support planes of GBSVM in Fig. \ref{fig:GBSVM} are constrained as follows:
	\begin{align}
		l_1^\prime:\quad\quad \frac {\omega}{\left\| \omega  \right\|} \cdot (c_i-x_i)\ge r_i, y_i=+1, \quad \label{equ:l'1o}\\
		l_2^\prime:\quad\quad \frac {\omega}{\left\| \omega  \right\|} \cdot (c_i-x_i)\le -r_i, y_i=-1.  \label{equ:l'2o}
	\end{align}
	Further, $l_1^\prime$ and $l_2^\prime$ can also be expressed as:
	\begin{align}
		l_1^\prime:\quad\quad \frac {\omega}{\left\| \omega  \right\|} \cdot (c_i-x_i)\ge y_i*r_i, y_i=+1,\label{equ:l'1}\\
	 l_2^\prime:\quad\quad \frac {\omega}{\left\| \omega  \right\|} \cdot (c_i-x_i)\le y_i*r_i, y_i=-1. \label{equ:l'2} 
	\end{align}
	Multiplying both sides of Eqs. (\ref{equ:l'1}) and (\ref{equ:l'2}) by $y_i$, and combining Eq. (\ref{equ:l12-n}), we combine them into one formula:
	\begin{equation}\label{equ:l'12}
		\begin{aligned}
			y_i\omega \cdot c_i-(y_i)^2+y_i b-\left\| \omega  \right\| r_i\ge 0.
		\end{aligned}
	\end{equation}
	Since $(y_i)^2=1$, Eq. (\ref{equ:l'12}) is changed as:
	\begin{equation}\label{equ:l'12-n}
		y_i\omega \cdot c_i+y_i b- \left\| \omega  \right\| r_i\ge 1.
	\end{equation}
	According to Eqs. (\ref{equ:l12-n}) and (\ref{equ:l'12-n}), the support planes $l_1^\prime$, $l_2^\prime$ and the decision plane $l^\prime$ are expressed as:
	\begin{equation}\label{equ:l'1-n}
		l_1^\prime:\quad\quad \omega \cdot c_i-\left\|\omega\right\| r_i+b=1,y_i=+1,
	\end{equation}
	\begin{equation}\label{equ:l'2-n}
		l_2^\prime:\quad\quad \omega \cdot c_i+\left\| \omega  \right\| r_i+b=-1,y_i=-1,
	\end{equation}
	\begin{equation}\label{equ:l'}
		l^\prime:\quad\quad \omega \cdot c_i+b=0.
	\end{equation}
	$l_1^\prime$ and $l_2^\prime$ are the tangent planes of $GB_1$ and $GB_2$ respectively as shown in Fig. \ref{fig:GBSVM}(a), so the interval between the two support planes, i.e., the $margin$ of GBSVM, can be expressed as:
	\begin{equation}\label{equ:margin}
		\begin{split}
			margin=\frac{((c_{1},-r_{1})-(c_{2},r_{2})) \cdot (w,||w||)}{||(w,||w||)||} \\=  \frac{((c_{1},-r_{1})-(c_{2},r_{2})) \cdot (w,||w||)}{\sqrt{2||w||^2}}
			\\= \frac{(c_{1}-c_{2})w-(r_1+r_2)||w||}{\sqrt{2}||w||}.
		\end{split}        	
	\end{equation}
	In addition, the support plane (\ref{equ:l'1-n}) minus (\ref{equ:l'2-n}), we get:
	\begin{equation}\label{equ:l1-l2}    
		l_1^\prime-l_2^\prime=\omega \cdot (c_1-c_2)-\left\| \omega  \right\|(r_1+r_2)=2.
	\end{equation}
	So, Eq. (\ref{equ:margin}) can be changed as:
	\begin{equation}\label{2}
		margin=\frac{\sqrt{2}}{\left\| \omega  \right\|}.
	\end{equation}
	In order to get the optimal decision plane, $margin$ must be maximized, which is consistent with the traditional SVM. Therefore, the objective function of GBSVM is:
	\begin{equation}\label{equ:oobj}
		\begin{aligned}
			\substack{\max\\w,b}\quad\quad &\frac{\sqrt{2}}{\left\| \omega  \right\|},\\
			s.t.\quad\quad &y_i(\omega \cdot c_i+b)-\left\| \omega  \right\| r_i\ge 1\text,\quad i=1,2,\cdots.
		\end{aligned}
	\end{equation}
	Eq. (\ref{equ:oobj}) is equivalent with the convex function:
	\begin{equation}\label{equ:oobj-convex}
		\begin{aligned}
			\substack{\min\\\omega,b}\quad\quad &\frac{1}{2} \left\| \omega  \right\|^2,\\
			s.t.\quad\quad &y_i(\omega \cdot c_i+b)-\left\| \omega  \right\| r_i\ge 1 \text,\quad i=1,2,\cdots.
		\end{aligned}
	\end{equation}

   To summarize, the strict derivation yields the convex quadratic programming model of GBSVM, which is expressed as the equation (\ref{equ:oobj-convex}). Notably, when the value of $r$ equals 0, the GBSVM model is equivalent to the traditional SVM. In this case, the granular-ball containing the sample becomes the finest size, and the equivalence of the two models confirms the correctness of the GBSVM model.
	\paragraph{The Dual Model of the Separable GBSVM}
	After introducing Lagrange multiplier $\alpha_i$ for the inequality constraint, the Lagrange function of Eq. (\ref{equ:oobj-convex}) can be expressed as:
	\begin{equation}\label{equ:lagrange}
		\begin{aligned}
			L(\omega,b,\alpha)=\frac{1}{2}\left\| \omega  \right\|^2-\sum\limits_{i=1}^{\textcolor{black}{m}} \alpha_i(y_i(\omega \cdot c_i+b)-\left\| \omega  \right\| r_i-1).
		\end{aligned}
	\end{equation}
	Let $L(\omega,b,\alpha)$ on $\omega$ and $b$ partial derivatives equal to 0, and we get:
	\begin{align}
	\frac{\partial L}{\partial \omega}&=\omega-\sum\limits_{i=1}^{\textcolor{black}{m}} \alpha_i y_i c_i+\sum\limits_{i=1}^{\textcolor{black}{m}}\alpha_i r_i\frac{\omega}{\left\| \omega  \right\|} =0;\label{equ:pa-w}\\
	\frac{\partial L}{\partial b}&=-\sum\limits_{i=1}^{\textcolor{black}{m}}\alpha_i y_i=0.\label{equ:pa-b}
\end{align}
	Then, Eq. (\ref{equ:pa-w}) can be changed as:
	\begin{equation}\label{equ:w}
		\omega=\frac{\left\| \omega  \right\| \sum\limits_{i=1}^{\textcolor{black}{m}}\alpha_i y_i c_i}{\left\| \omega  \right\| +\sum\limits_{i=1}^{\textcolor{black}{m}}\alpha_i r_i}.
	\end{equation}
    In order to get an expression for $\omega$, square both sides of Eq. (\ref{equ:w}). We can get:
	\begin{equation}\label{equ:square}
		\left( \sum\limits_{i=1}^{\textcolor{black}{m}}\alpha_i y_i c_i \right)^2=\left( \left\| \omega  \right\| +\sum\limits_{i=1}^{\textcolor{black}{m}}\alpha_i r_i\right) ^2.
	\end{equation}
	The square root of Eq. (\ref{equ:square}) is:
	\begin{equation}\label{equ:square-root}
		\begin{aligned}
			\left\| \sum\limits_{i=1}^{{\textcolor{black}{m}}}\alpha_i y_i c_i \right\|=\left( \left\| \omega  \right\|+\sum\limits_{i=1}^{{\textcolor{black}{m}}}\alpha_i r_i \right).
		\end{aligned}
\end{equation}
	Since $\left\| \omega  \right\|\ge 0$, $\alpha_i\ge 0$ and $r_i\ge 0$, Eq. (\ref{equ:square-root}) is changed as:
	\begin{equation}\label{equ:s'-root}	
		\begin{aligned}
			\left\| \omega  \right\|= \left\| \sum\limits_{i=1}^{\textcolor{black}{m}}\alpha_i y_i c_i \right\|-\sum\limits_{i=1}^{\textcolor{black}{m}}\alpha_i r_i  .
		\end{aligned}
\end{equation}
	According to Eqs. (\ref{equ:s'-root}) and (\ref{equ:w}), $\omega$ can be rewritten as:  
	\begin{equation}\label{equ:w'}	
		\omega=\frac{\left( \left\| \sum\limits_{i=1}^{\textcolor{black}{m}}\alpha_i y_i c_i \right\|-\sum\limits_{i=1}^{\textcolor{black}{m}}\alpha_i r_i\right) \sum\limits_{i=1}^{\textcolor{black}{m}}\alpha_i y_i c_i}{\left\| \sum\limits_{i=1}^{\textcolor{black}{m}}\alpha_i y_i c_i \right\|} = \frac{(\|A\|-B)A}{\|A\|},
	\end{equation}
	where $A = \sum\limits_{i=1}^{\textcolor{black}{m}}\alpha_i y_i c_i$ and $B = \sum\limits_{i=1}^{\textcolor{black}{m}}\alpha_i r_i$. According to Eq. (\ref{equ:w'}), we have:
	\begin{equation}\label{equ:w''}
	\setlength{\abovedisplayskip}{0.5pt}
		\|\omega\|= \|A\|-B.
	 \setlength{\belowdisplayskip}{0.5pt}
	\end{equation}
	\textcolor{black}{Putting} Eqs. (\ref{equ:pa-w}) and (\ref{equ:w'}) into Eq. (\ref{equ:lagrange}), that is, all terms about $\omega$ are replaced by that of $\alpha_i$, we have: 
	\begin{equation}
			\setlength{\abovedisplayskip}{0.5pt}
		\begin{aligned}
			&\frac{1}{2}\left\| \omega  \right\|^2-\sum\limits_{i=1}^{\textcolor{black}{m}}\alpha_i(y_i(\omega \cdot c_i+b)-\left\| \omega  \right\| r_i-1) \\ &= \frac{1}{2}\left\| \omega  \right\|^2 - \sum\limits_{i=1}^{\textcolor{black}{m}}\alpha_i y_i c_i \omega - \sum\limits_{i=1}^{\textcolor{black}{m}}\alpha_i y_i b + \sum\limits_{i=1}^{\textcolor{black}{m}}\alpha_i r_i \left\| \omega  \right\| + \sum\limits_{i=1}^{\textcolor{black}{m}} \alpha_i  \\&= \frac{1}{2} (\|A\|-B)^2 -A \frac{(\|A\|-B)A}{\|A\|} + B (\|A\|-B) + \sum\limits_{i=1}^{\textcolor{black}{m}}\alpha_i  \\ &= - \frac{1}{2} A^2 + \frac{1}{2} B^2 + (\|A\|-B)B + \sum\limits_{i=1}^{\textcolor{black}{m}}\alpha_i \\ & = -\frac{1}{2}A^2 - \frac{1}{2}B^2 + \|A\|B+\sum\limits_{i=1}^{\textcolor{black}{m}}\alpha_i.
		\end{aligned}
		 \setlength{\belowdisplayskip}{0.5pt}
	\end{equation}
	
	The dual function of Eq. (\ref{equ:oobj-convex}) is:
	\begin{equation}\label{7}
			\setlength{\abovedisplayskip}{0.5pt}
		\begin{aligned}
			\substack{\max\\\alpha}\quad\quad &-\frac{1}{2}A^2 - \frac{1}{2}B^2 + \|A\|B+\sum\limits_{i=1}^{\textcolor{black}{m}}\alpha_i,\\
			s.t.\quad\quad &\sum\limits_{i=1}^{\textcolor{black}{m}}\alpha_i y_i=0 \text,\quad i=1,2,\cdots,
		\end{aligned}
	\end{equation}
	where $A = \sum\limits_{i=1}^{\textcolor{black}{m}}\alpha_i y_i c_i$ and $B = \sum\limits_{i=1}^{\textcolor{black}{m}}\alpha_i r_i$. Eq. (\ref{7}) can be transformed to the same function with the original model of SVM as follows:
	\begin{equation}\label{equ:finalooj2}
		\begin{aligned}
			\substack{\max\\\alpha}\quad\quad &-\frac{1}{2} \left\|w\right\|^2+\sum\limits_{i=1}^{\textcolor{black}{m}}\alpha_i,\\
			s.t.\quad\quad &\sum\limits_{i=1}^{\textcolor{black}{m}}\alpha_i y_i=0 \text,\quad i=1,2,\cdots,
		\end{aligned}
			 \setlength{\belowdisplayskip}{0.5pt}
	\end{equation}
	where $w$ is as shown in Eq. (\ref{equ:w'}).
	
	\subsubsection{The Inseparable GBSVM}
	
	\paragraph{The Original Model of the Inseparable GBSVM}
	In the separable GBSVM, all the support granular-balls must satisfy the constraints. However, some granular-balls can not meet the constraints in most cases, so we introduce the slack variable $\xi$ and penalty coefficient $C$, whose roles are the same as SVM. Thus, the inseparable GBSVM model is expressed as:
	\begin{equation}\label{equ:OObj-i}
		\begin{aligned}
			\substack{\min\\\omega,b,\xi_i}\quad\quad &\frac{1}{2} \left\| \omega  \right\|^2+C\sum\limits_{i=1}^{\textcolor{black}{m}}\xi_i,\\
			s.t.\quad\quad &y_i(\omega c_i+b)-\left\| \omega  \right\| r_i\ge 1-\xi_i \text,\quad i=1,2,\cdots\\
			\quad\quad &\xi_i\ge 0.
		\end{aligned}
	\end{equation}
	\paragraph{The Dual Model of the Inseparable GBSVM}
	After introducing the Lagrange multiplier $\alpha_i$ for the inequality constraint, the Lagrange function of Eq. (\ref{equ:OObj-i}) can be expressed as:
	\begin{small}
		\begin{equation}\label{equ:lagerange-i}
						\setlength{\abovedisplayskip}{0.5pt}
			\begin{aligned}
				L(\omega,b,\xi,\alpha,\mu)=&\frac{1}{2}\left\| \omega \right\|^2 -\sum\limits_{i=1}^{\textcolor{black}{m}}\alpha_i(y_i(\omega c_i+b)-\left\| \omega  \right\| r_i-1+\xi_i)\\
				&+C\sum\limits_{i=1}^{\textcolor{black}{m}}\xi_i-\sum\limits_{i=1}^{\textcolor{black}{m}}\mu_i\xi_i.
			\end{aligned}	
			 \setlength{\belowdisplayskip}{0.5pt}
		\end{equation}
	\end{small}
	Let $L(\omega,b,\xi,\alpha,\mu)$ on $\omega$, $b$ and $\xi$ partial derivatives equal to 0, we get:
	\begin{equation}\label{equ:pd-w}
					\setlength{\abovedisplayskip}{0.5pt}
		\frac{\partial L}{\partial \omega}=\omega-\sum\limits_{i=1}^{\textcolor{black}{m}}\alpha_i y_i c_i+\sum\limits_{i=1}^{\textcolor{black}{m}}\alpha_i r_i\frac{\omega}{\left\| \omega  \right\|} =0,
   \setlength{\belowdisplayskip}{0.5pt}
	\end{equation}
	\begin{equation}\label{equ:pd-b}
		\frac{\partial L}{\partial b}=-\sum\limits_{i=1}^{\textcolor{black}{m}}\alpha_i y_i=0,
	\end{equation}
	and
	\begin{equation}\label{equ:pd-xi}
		\frac{\partial L}{\partial \xi}=C-\alpha_i-\mu_i =0.
	\end{equation}
	Putting Eqs. (\ref{equ:pd-w}), (\ref{equ:pd-b}) and (\ref{equ:pd-xi}) into Eq. (\ref{equ:lagerange-i}), the dual model of the inseparable GBSVM is:
	\begin{equation}\label{equ:insobj}
 	\setlength{\abovedisplayskip}{0.5pt}
		\begin{aligned}
			\substack{\max\\\alpha}\quad\quad &-\frac{1}{2}A^2 - \frac{1}{2}B^2 +  \|A\|B +\sum\limits_{i=1}^{\textcolor{black}{m}}\alpha_i,\\
			s.t.\quad\quad &\sum\limits_{i=1}^{{\textcolor{black}{m}}}\alpha_i y_i=0,\\
			\quad\quad &0\le \alpha_i \le C \text,\quad i=1,2,\cdots,
		\end{aligned}
	\setlength{\belowdisplayskip}{0.5pt}
	\end{equation}
	where $A = \sum\limits_{i=1}^{{\textcolor{black}{m}}}\alpha_i y_i c_i$ and $B = \sum\limits_{i=1}^{{\textcolor{black}{m}}}\alpha_i r_i$. 
	
	Eq. (\ref{equ:insobj}) can be transformed to the same function with the original model of SVM as follows:
	\begin{equation}\label{equ:finalooj}
		\begin{aligned}
			\substack{\max\\\alpha}\quad\quad &-\frac{1}{2} \left\| w \right\|^2+\sum\limits_{i=1}^{{\textcolor{black}{m}}}\alpha_i,\\
			s.t.\quad\quad &\sum\limits_{i=1}^{{\textcolor{black}{m}}}\alpha_i y_i=0,\\
			\quad\quad &0\le \alpha_i \le C \text,\quad i=1,2,\cdots.
		\end{aligned}
	\end{equation}
	where $w$ is shown in Eq. (\ref{equ:w'}).
	
	\textbf{Comment:}
 	 Up to this point, we have derived the dual models of separable SVM and inseparable SVM in linear classification. From another perspective, the main advantage of SVM is in processing linear classification, in which SVM can find a global optimal solution; so, it can achieve a higher accuracy than other classifiers in linear classification. Therefore, GBSVM is mainly to deal with data classification with label noise. The linear separable GBSVM model (\ref{equ:oobj-convex}) and the inseparable model (\ref{equ:OObj-i}) show that when the radius is set to $0$, the GBSVM models are equivalent to the SVM. Moreover, the forms of $w$ (\ref{equ:w'}) and the dual model (\ref{7}) and (\ref{equ:insobj}) align with traditional SVM when the radius is set to $0$, which is a concise and noteworthy conclusion. Furthermore, their respective dual models (equations (\ref{equ:finalooj2}) and (\ref{equ:finalooj})) are also consistent with the traditional model. Consequently, the maximum time complexity for traditional SVM is $O(\textcolor{black}{n}^3)$ \cite{burges1998tutorial}, where $\textcolor{black}{n}$ denotes the number of sample points. The maximum time complexity for GBSVM is $O(\textcolor{black}{m}^3)$, where $\textcolor{black}{m}$ corresponds to the number of granular-balls. These time complexities are consistent with each other. Using the granular-balls instead of sample points as input greatly reduces the number of input samples, i.e. $\textcolor{black}{m}$ being much smaller than $\textcolor{black}{n}$. Consequently, the training efficiency of GBSVM is higher than that of the traditional SVM. For non-linear classification, the kernel function needs to be introduced, which we will address in the next subsection.
 	   	\vspace{-0.2cm}
  \subsection{Non-linear Granular-ball Support Vector Machine}	
  	\vspace{-0.4cm}
  	\begin{figure}[ht]
  		\setlength{\abovecaptionskip}{-0.1cm}
  	\centering
  	\includegraphics[width = 0.5\textwidth]{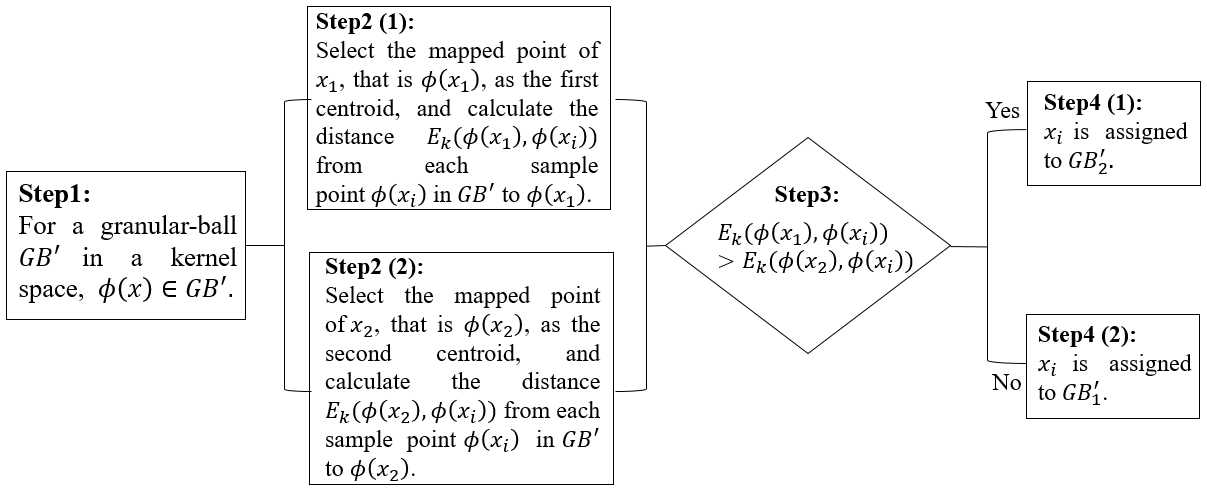}\\
  	\caption{\textcolor{black}{The split process of granular-balls in the kernel space.}} 
  	\label{fig:kernel-ball}
  \end{figure}
  	\vspace{-0.2cm}
		When it is impossible to find a hyperplane in the original dimension to separate the data into two categories, the kernel function is used to perform a dimension-raising operation and locate the hyperplane in a higher dimensional space to divide the two categories.
	 \textcolor{black}{Therefore, nonlinear GBSVM needs to be trained on granular-balls in high dimensional space. Assuming that the center and radius of a granular-ball in the high-dimensional space are $c^\prime$ and $r^\prime$ respectively, the nonlinear GBSVM model is as follows:
		\begin{equation}\label{nGBSVM}
			\begin{aligned}
				&	\substack{\max\\\alpha}\quad  -\frac{1}{2}\sum_{i=1}^{m} \sum_{j=1}^{m} \alpha_i \alpha_j y_i y_j c_i^\prime c_j^\prime - \frac{1}{2}\sum_{i=1}^{m} \sum_{j=1}^{m} \alpha_i \alpha_j r_i^\prime r_j^\prime  \quad \quad \quad \quad \quad \\ &+  \left\|  \sum_{i=1}^{m} \alpha_i y_i c_i^\prime \right\|   \ \sum_{i=1}^{m} \alpha_i r_i^\prime  +\sum\limits_{i=1}^{m}\alpha_i,\\
				&	s.t.\quad\quad \sum\limits_{i=1}^{m}\alpha_i y_i=0 \text,\quad i=1,2,\cdots.
			\end{aligned}
			\setlength{\belowdisplayskip}{0.5pt}		
		\end{equation}
		It is a challenge to calculate granular-balls in high dimensional space. It is unreasonable to directly map the granular-ball $GB_i(c_i,r_i)$ generated in the original space into a high dimensional space, because the radius cannot be directly mapped into a high dimensional space. Another method is to generate granular-balls in the original space and map the points in each ball into the kernel space to generate new granular-balls. Although the mapped granular-balls contain the same sample points as the original balls, it cannot accurately describe the data set because the mapped data distribution has changed. Generating granular-balls in kernel space is a better solution. In this method, the key problem is that the mapping function $\phi(x)$ is unknown, so the split process of granular-balls is not explicit. However, we can learn from the kernel-kmeans method \cite{dhillon2004kernel} to generate granular-balls in an implicit way. The splitting process is shown in Fig. \ref{fig:kernel-ball}. Although the mapping function $\phi(x)$ in steps 1-2 is unknown, steps 4-5 can still be performed by calculating the distance $E_k$ as follows: 
			\begin{equation} 
      \setlength{\abovedisplayskip}{0.6pt}
			\begin{split}
				E_k (\phi (x_1),\phi (x_i) ) &=  \left\|  \phi (x_1)-\phi (x_i)\right\| \\ & = \sqrt{K_{x_1,x_1}-2K_{x_1,x_i} +K_{x_i,x_i}},
			\end{split}	
			\end{equation}	
		where $K$ denotes the kernel function. In addition, the center and radius of a mapped granular-ball can be calculated as follows:
		\begin{equation}
       \setlength{\abovedisplayskip}{0.5pt}
		\begin{split}
			& \ \ \ \ \ \ \ \ \ \ \ \ \ \ \ \ \ \ c^\prime=\frac{1}{k}\sum_{i=1}^{k}\phi\left(x_i\right), \\
			&r^\prime=\frac{1}{k}\sum_{i=1}^{k}  \left\| \phi\left(x_i\right)-c^\prime \right\| = \frac{1}{k}\sum_{i=1}^{k} \sqrt{\left( \phi \left( x_{i}\right) - c^\prime \right) ^{2}} \\ & \ \  = \frac{1}{k}\sum_{i=1}^{k} \sqrt{ K_{x_i,x_i} - \frac{1}{k}\sum_{j=1}^{k} K_{x_i,x_j} + C_{c^\prime,c^\prime}},
		\end{split}
			\setlength{\belowdisplayskip}{2.5pt} 	
	\end{equation}	
where $k$ denotes the number of samples in a granular-ball. The inner product $C_{c_i^\prime,c_j^\prime}$ of centers $c_i^\prime$ and $c_j^\prime$ can be calculated as
\begin{equation}
 \setlength{\abovedisplayskip}{0.5pt}
	\begin{split}
		&C_{c_i^\prime,c_j^\prime}=\left\langle {c}_i^\prime,c_j^\prime\right\rangle  =\left(c_i^\prime\right)^T\left(c_j^\prime\right) =\frac{1}{k_1 k_2}\sum_{ii=1}^{k_1}{\phi\left(x_{ii}\right)^T}\sum_{jj=1}^{k_2}{\phi\left(x_{jj}\right)}\\
		&= \frac{1}{k_1 k_2} \sum_{ii=1}^{k_1}\sum_{jj=1}^{k_2}{\phi\left(x_{ii}\right)^T \phi\left(x_{jj}\right)} = \frac{1}{k_1 k_2} \sum_{ii=1}^{k_1}\sum_{jj=1}^{k_2}K_{x_{ii},x_{jj}}.
	\end{split}
	\setlength{\belowdisplayskip}{0.5pt}
\end{equation}	
 The third term of (\ref{nGBSVM}) can be simplified as follows:
 \begin{equation} 
 \setlength{\abovedisplayskip}{0.5pt}
 			\setlength{\abovedisplayskip}{0.5pt}
 	\begin{aligned}
 		&\left\|\sum_{i=1}^{m} \alpha_i y_i c_i^\prime \right\| \sum_{i=1}^{m} \alpha_i r_i^\prime = \sqrt{\left(\sum_{i=1}^{m} \alpha_i y_i c_i^\prime \right)^2} \ * \ \sum_{i=1}^{m} \alpha_i r_i^\prime . 
 	\end{aligned}
  \setlength{\belowdisplayskip}{0.5pt}
 \end{equation}
	Therefore, the separable GBSVM dual model (\ref{nGBSVM}) with a kernel function is derived as:
\begin{equation}\label{n7}
				\setlength{\abovedisplayskip}{0.5pt}
	\begin{aligned}
		&	\substack{\max\\\alpha}\quad  -\frac{1}{2}\sum_{i=1}^{m} \sum_{j=1}^{m} \alpha_i \alpha_j y_i y_j C_{c_i^\prime,c_j^\prime} - \frac{1}{2}\sum_{i=1}^{m} \sum_{j=1}^{m} \alpha_i \alpha_j r_i^\prime r_j^\prime  \quad \quad \quad \quad \quad \\ &+ \sqrt{\left(\sum_{i=1}^{m} \alpha_i y_i c_i^\prime \right)^2 } * \ \sum_{i=1}^{m} \alpha_i r_i^\prime  +\sum\limits_{i=1}^{m}\alpha_i,\\
		&	s.t.\quad\quad \sum\limits_{i=1}^{m}\alpha_i y_i=0 \text,\quad i=1,2,\cdots.
	\end{aligned}
	\setlength{\belowdisplayskip}{0.5pt}
\end{equation}
According to (\ref{equ:insobj}), the inseparable GBSVM dual model with the kernel function is as follows:
\begin{equation}\label{n8}
 \setlength{\abovedisplayskip}{0.5pt}
	\begin{aligned}
		&	\substack{\max\\\alpha}\quad  -\frac{1}{2}\sum_{i=1}^{m} \sum_{j=1}^{m} \alpha_i \alpha_j y_i y_j C_{c_i^\prime,c_j^\prime} - \frac{1}{2}\sum_{i=1}^{m} \sum_{j=1}^{m} \alpha_i \alpha_j r_i^\prime r_j^\prime  \quad \quad \quad \quad \quad \\ &+ \sqrt{\left(\sum_{i=1}^{m} \alpha_i y_i c_i^\prime \right)^2 } * \ \sum_{i=1}^{m} \alpha_i r_i^\prime  +\sum\limits_{i=1}^{m}\alpha_i,\\
		&	s.t.\quad\quad \sum\limits_{i=1}^{m}\alpha_i y_i=0 \text,\quad i=1,2,\cdots,\\
		&	\quad \quad \quad \quad 0\le \alpha_i \le C \text,\quad i=1,2,\cdots.
	\end{aligned}
	\setlength{\belowdisplayskip}{0.5pt}
\end{equation}   
	}	
  The presence of $|| \cdot ||$ in the model poses some difficulty in solving it. Therefore, we use two algorithms, PSO and SMO, to solve the model in the algorithm design chapter.

	\subsection{Algorithm Design}
	
	\subsubsection{Particle swarm optimization algorithm for GBSVM}  
	The particle swarm optimization algorithm (PSO) \cite{flake2002efficient} is used to solve the model, and the algorithm is shown \textcolor{black}{in the file ``Supplementary\_Material.pdf''}. In Step 5, when a $\alpha_i$ is smaller than zero, it will be set to zero. When $\sum\limits_{i=1}^{m}\alpha_i y_i>0$, all the $\alpha_i$ corresponded to those data points labeled with `+1' are multiplied with a coefficient between 0 and 1, and those $\alpha_i$ corresponded to those data points labeled with `-1' are multiplied with the inverse of that coefficient. As a result, $\sum\limits_{i=1}^{m}\alpha_i y_i$ can be decreased to zero. When $\sum\limits_{i=1}^{m}\alpha_i y_i<0$, all the $\alpha_i$ corresponded to those data points labeled with `-1' are multiplied with a coefficient between 0 and 1, and those $\alpha_i$ corresponded to those data points labeled with `+1' are multiplied with the inverse of that coefficient. As a result, $\sum\limits_{i=1}^{m}\alpha_i y_i$ can be decreased to zero. In this process, all the Lagrange coefficients are optimized in an approximate way in which both the fitness function and constraints are considered at the same time. The process can be described as follows:
	\begin{equation} \label{equ:coeff1}
		\delta\sum\limits_{y_i=+1}\alpha_i y_i+\frac{1}{\delta}\sum\limits_{y_j=-1}\alpha_j y_j = 0,
	\end{equation}
	where $\delta$ is the coefficient.
	From (\ref{equ:coeff1}), we have:
	\begin{equation} \label{equ:coeff2}
		\delta=\sqrt{\frac{-\sum\limits_{y_j=-1}\alpha_j y_j}{\sum\limits_{y_i=+1}\alpha_i y_i}}.
	\end{equation}


Although the PSO algorithm can directly solve the dual model, it is a heuristic algorithm with a search process for each optimization, leading to lower time efficiency than the SMO algorithm.

	\subsubsection{Sequential minimal optimization algorithm for GBSVM}
		SVM is usually solved with pairwise problems, which has two advantages: first, there are only $N$ variables, where $N$ is the number of samples in the training set. The number of variables in the original problem is the same as the number of features in the sample points, which is more difficult to solve when the sample features are very large. Second, it is convenient to introduce kernel functions to solve nonlinear SVM. The common algorithm for solving pairwise problems is sequential minimal optimization(SMO) algorithm \cite{platt1998sequential}, which is based on the idea that in each optimization step, two of the many parameters $ \alpha_1,\alpha_ 2$ are selected as the ``true parameters" and the rest of the parameters are considered as constants. For the dual model of separable and inseparable GBSVM, it is difficult to use the SMO algorithm to optimize the objective function to obtain the optimal solution. The reason is that the objective function contains the norm term $\left\| A \right\| $, which cannot extract the parameters $\alpha_1,\alpha_2$. The original objective function can be obtained from the dual models (\ref{7}) and (\ref{equ:insobj}) as:
		\begin{equation}\label{mu1}
			\begin{aligned}
				f&=\substack{\max\\\alpha} -\frac{1}{2}A^2 - \frac{1}{2}B^2 +  \|A\|B +\sum\limits_{i=1}^{\textcolor{black}{m}}\alpha_i \\
				&= \substack{\min\\\alpha} \frac{1}{2}A^2 + \frac{1}{2}B^2 -  \|A\|B -\sum\limits_{i=1}^{\textcolor{black}{m}}\alpha_i \\
				&=\substack{\min\\\alpha}\frac{1}{2}(\|A\|-B)^2 - \sum\limits_{i=1}^{\textcolor{black}{m}}\alpha_i.
			\end{aligned}
		\end{equation} 
		Since $\left\| w \right\|\ge 0 $, then $ \left\| A \right\| \ge B$, we have the following inequality:
		\begin{equation}
			\begin{aligned}
				\frac{1}{2}(\|A\|-B)(\|A\|-B) - \sum\limits_{i=1}^{\textcolor{black}{m}}\alpha_i & \le 	\frac{1}{2}(\|A\|^2 - B^2)  - \sum\limits_{i=1}^{\textcolor{black}{m}}\alpha_i \\ & \le \frac{1}{2} \|A\|^2 - \sum\limits_{i=1}^{\textcolor{black}{m}}\alpha_i.
			\end{aligned}
		\end{equation}
		If we consider using $ \frac{1}{2} \|A\|^2 - \sum\limits_{i=1}^{\textcolor{black}{m}}\alpha_i $ as an approximate function, the error is large and the effect of radius is not considered. Therefore, we consider $\frac{1}{2}(\|A\|^2 - B^2)  - \sum\limits_{i=1}^{\textcolor{black}{m}}\alpha_i $ as an approximate function and introduce a weight parameter $\lambda$ to reduce the loss. The following functions are considered as the approximate functions:
		\begin{equation}
			\frac{1}{2} (\|A\|^2 - \lambda  B^2 )  - \sum\limits_{i=1}^{\textcolor{black}{m}}\alpha_i .
			\setlength{\belowdisplayskip}{3pt}
		\end{equation}	
		Here $\lambda$ is the weight parameter, which can greatly reduce the loss. It can be obtained by the following equation:
		\begin{equation}\label{js}
			(\|A\|-B)^2 = \|A\|^2 - \lambda  B^2. 
		\end{equation}
		In the iterative process, the parameter value is adaptive and will accordingly correspond to the iteration. We set the $\lambda^t$ of the $t$-th step by using the $\alpha^{t-1}_i$, as follows:
		\begin{equation}\label{lam}
			\lambda^t =  \frac{2\|A\| - B }{B}= \frac{2\left\| \sum\limits_{i=1}^{m}\alpha^{t-1}_i y_i c_i \right\|-\sum\limits_{i=1}^{\textcolor{black}{m}}\alpha^{t-1}_i r_i}{\sum\limits_{i=1}^{\textcolor{black}{m}}\alpha^{t-1}_i r_i}.
		\end{equation}
		In fact, both the approximation function and the objective function are convex and have unique minimal solutions. The loss between the approximation function and the objective function will become smaller and smaller as the $\alpha$ iterations, and the values of the two functions will be infinitely close when the algorithm finally converges. Therefore, the solution of GBSVM can be approximated by the approximate objective function.
		
	\textbf{Linear GBSVM:}
    	Therefore, according to the approximate equation (\ref{js}), the linear objective function of optimization is:
		\begin{equation}\label{jinsi}
			\begin{aligned}
				&\frac{1}{2} (\|A\|^2 - \lambda  B^2 )  - \sum\limits_{i=1}^{\textcolor{black}{m}}\alpha_i \\  &= \frac{1}{2} \sum\limits_{i=1}^{\textcolor{black}{m}} \sum\limits_{j=1}^{\textcolor{black}{m}} \alpha_i \alpha_j y_i y_j c^{T}_i c_{j} - \frac{\lambda}{2} \sum\limits_{i=1}^{\textcolor{black}{m}} \sum\limits_{j=1}^{\textcolor{black}{m}} \alpha_i \alpha_j r_i r_j - \sum\limits_{i=1}^{\textcolor{black}{m}}\alpha_i.
			\end{aligned}
		\end{equation}	
		According to the SMO algorithm, only two variables $\alpha_1,\alpha_2$ are optimized at a time, and other variables $\alpha_3,\alpha_4,...,\alpha_n$ are treated as constants. All constants are removed from the objective function into the following formula:
		\begin{equation}\label {a1}
			\begin{aligned}
				\substack{\min\\\alpha_1,\alpha_2} \ \ & \frac{1}{2} \alpha^2_1 c^{T}_1 c_{1} + \frac{1}{2} \alpha^2_2 c^{T}_2 c_{2} + \alpha_1 \alpha_2 y_1 y_2 c^{T}_1 c_{2} \\ &+  \sum\limits_{i=3}^{\textcolor{black}{m}} \alpha_1 y_1 \alpha_i y_i c^{T}_1 c_{i} + \sum\limits_{i=3}^{\textcolor{black}{m}} \alpha_2 y_2 \alpha_i y_i c^{T}_2 c_{i} - \frac{\lambda}{2} \alpha^2_1 r^2_1 \\ &- \frac{\lambda}{2} \alpha^2_2 r^2_2 - \lambda \alpha_1 \alpha_2  r_1 r_2 - \lambda \sum\limits_{i=3}^{\textcolor{black}{m}} \alpha_1 \alpha_i  r_1 r_i \\ &- \lambda \sum\limits_{i=3}^{\textcolor{black}{m}} \alpha_2 \alpha_i  r_2 r_i - \alpha_1 - \alpha_2 .
			\end{aligned}
		\end{equation}
		The constraints are 
		\begin{equation}\label{con}
			\begin{aligned}
				\alpha_1 y_1 + \alpha_2 y_2 = - \sum\limits_{i=3}^{\textcolor{black}{m}} \alpha_i y_i = \zeta ; 0 < \alpha_i < C.
			\end{aligned}
		\end{equation}
		Bringing $\alpha_1=  (\zeta - \alpha_2 y_2)y_1 $ into Eq. (\ref{a1}), the quadratic function with respect to a single variable $\alpha_2$ can be obtained as
		\begin{equation}\label{er}
			\begin{aligned}
				 \frac{1}{2} &(\zeta - \alpha_2 y_2)^2 c^{T}_1 c_{1} + \frac{1}{2} \alpha^2_2 c^{T}_2 c_{2} + (\zeta  - \alpha_2 y_2)y_1 \alpha_2 y_2 c^{T}_1 c_{2} \\ & + \sum\limits_{i=3}^{\textcolor{black}{m}} (\zeta - \alpha_2 y_2) \alpha_i y_i c^{T}_1 c_{i} + \sum\limits_{i=3}^{\textcolor{black}{m}} \alpha_2 y_2 \alpha_i y_i c^{T}_2 c_{i}  \\ &- \frac{\lambda}{2} (\zeta - \alpha_2 y_2)^2  r^2_1 - \frac{\lambda}{2} \alpha^2_2 r^2_2 - \lambda (\zeta - \alpha_2 y_2)y_1 \alpha_2  r_1 r_2  \\ &- \lambda \sum\limits_{i=3}^{\textcolor{black}{m}} (\zeta - \alpha_2 y_2)y_1 \alpha_i  r_1 r_i- \lambda \sum\limits_{i=3}^{\textcolor{black}{m}} \alpha_2 \alpha_i  r_2 r_i \\ & - (\zeta - \alpha_2 y_2)y_1 - \alpha_2 .
			\end{aligned}
		\end{equation}
		By setting its derivative of $\alpha_2$ to 0, we get
		\begin{equation}\label{co1}
			\begin{aligned}
				&(c^{T}_1 c_{1} + c^{T}_2 c_{2} - 2 c^{T}_1 c_{2} - \lambda r^2_1 - \lambda r^2_2 +  2 \lambda y_1 y_2 r_1 r_2) \alpha_2 \\ &= \zeta y_2 c^{T}_1 c_{1}-\zeta y_2 c^{T}_1 c_{2} + \sum\limits_{i=3}^{\textcolor{black}{m}} y_2 \alpha_i y_i c^{T}_1 c_{i}-\sum\limits_{i=3}^{\textcolor{black}{m}} y_2 \alpha_i y_i c^{T}_2 c_{i} \\ &- \lambda \zeta y_2 r^2_1 + \lambda \zeta y_1 r_1 r_2 - \lambda y_1 y_2 \sum\limits_{i=3}^{\textcolor{black}{m}} r_1 r_i \\ &+ \lambda \sum\limits_{i=3}^{\textcolor{black}{m}} r_2 r_i - y_1 y_2 +1.
			\end{aligned}
		\end{equation}
		In addition, the values of $\alpha_1$ and $\alpha_2$ in the last iteration denote $\alpha^{old}_1$ and $\alpha^{old}_2$ respectively,  and we have
		\begin{equation}\label{con2}
			\zeta=\alpha^{old}_1 y_1 + \alpha^{old}_2 y_2.
		\end{equation}
		Putting Eqs. (\ref{con2}) into (\ref{co1}), $\alpha_2$ can be updated as follows:
		\begin{equation}\label{con3}
			\alpha^{new,unc}_2 = \frac{Q_1}{ c^{T}_1 c_{1} + c^{T}_2 c_{2} - 2 c^{T}_1 c_{2} - \lambda r^2_1 - \lambda r^2_2 +  2 \lambda y_1 y_2 r_1 r_2 },
		\end{equation}	
		where 
		\begin{equation}
			\begin{aligned}
				Q_1&=\zeta y_2 c^{T}_1 c_{1}-\zeta y_2 c^{T}_1 c_{2} + \sum\limits_{i=3}^{\textcolor{black}{m}} y_2 \alpha_i y_i c^{T}_1 c_{i}-\sum\limits_{i=3}^{\textcolor{black}{m}} y_2 \alpha_i y_i c^{T}_2 c_{i} \\ &- \lambda \zeta y_2 r^2_1 + \lambda \zeta y_1 r_1 r_2 - \lambda y_1 y_2 \sum\limits_{i=3}^{\textcolor{black}{m}} r_1 r_i \\ &+ \lambda \sum\limits_{i=3}^{\textcolor{black}{m}} r_2 r_i - y_1 y_2 +1.
			\end{aligned}
		\end{equation}	
		The final $\alpha^{new}_2$ satisfies condition (\ref{con}), that is:
		\begin{equation}\label{con4}
			\alpha^{new}_2=\begin{array}{l} 
				\left\{\begin{matrix} 
					H ,    \ \alpha^{new,unc}_2> H; \\ 
					\alpha^{new,unc}_2, \  L \le  \alpha^{new}_2 \le H; \\ 
					L , \    \alpha^{new,unc}_2< L,
				\end{matrix}\right.    
			\end{array} 
		\end{equation}	
		\textcolor{black}{where $L$ and $H$ are the boundary values of $\alpha^{new}_2$. When $y_1 \neq y_2$, there are} 
		\begin{equation}
			L=\max(0,\alpha^{old}_2-\alpha^{old}_1),H=\min(C,C+\alpha^{old}_2-\alpha^{old}_1).
		\end{equation}	
	\textcolor{black}{When $y_1=y_2$, there are } 
		\begin{equation}
			L=\max(0,\alpha^{old}_2 + \alpha^{old}_1-C),H=\min(C,\alpha^{old}_2+\alpha^{old}_1).
		\end{equation}	
\textcolor{black}{The derivation of $L$ and $H$ is similar to that of traditional SVM. We can get $\alpha^{new}_1$ according to $\alpha^{new}_2$ as
	\begin{equation}\label{con5}
		 \alpha^{new}_1 = \alpha^{old}_1 +y_1 y_2 (\alpha^{old}_2 - \alpha^{new}_2 ).
	\end{equation}	}
	In summary, the specific SMO algorithm flow is shown in Algorithm \ref{Algorithm1}.

	\textbf{Nonlinear GBSVM:}	
    For the nonlinear GBSVM model (\ref{n8}), the kernel function is introduced, and the objective function can be approximated according to the formula (\ref{jinsi}):
    \textcolor{black}{\begin{equation}\label{jinsin}
    	\begin{aligned}
    		&\frac{1}{2} (\|A\|^2 - \lambda  B^2 )  - \sum\limits_{i=1}^{m}\alpha_i \\ &= \frac{1}{2} \sum\limits_{i=1}^{m} \sum\limits_{j=1}^{m} \alpha_i \alpha_j y_i y_j C_{c_i^\prime, c_j^\prime} - \frac{\lambda}{2} \sum\limits_{i=1}^{m} \sum\limits_{j=1}^{m} \alpha_i \alpha_j  r_i^\prime r_j^\prime  - \sum\limits_{i=1}^{m}\alpha_i.
    	\end{aligned}
    \end{equation} 
Then, according to the formula (\ref{a1})-(\ref{con3}) in the linear GBSVM, ${\alpha^{new,unc}_2}^ \prime$ can be calculated as follows:
\begin{small}
	\begin{equation}\label{con23}
		\begin{aligned}
		 &{\alpha^{new,unc}_2}^ \prime = \\
		 &\frac{Q_2}{C_{c_1^\prime,c_1^\prime} + C_{c_2^\prime,c_2^\prime}-2C_{c_1^\prime,c_2^\prime} - \lambda  r_1^\prime r_1^\prime  - \lambda  r_2^\prime r_2^\prime  +  2 \lambda y_1 y_2 r_1^\prime r_2^\prime } ,
	\end{aligned}
	\end{equation}
\end{small}
	where 
\begin{equation}
		\begin{aligned}
		Q_2&=\zeta y_2 C_{c_1^\prime,c_1^\prime}-\zeta y_2 C_{c_1^\prime,c_2^\prime}+\sum\limits_{i=3}^{m} y_2 \alpha_i y_i C_{c_1^\prime,c_i^\prime} \\ &-\sum\limits_{i=3}^{m} y_2 \alpha_i y_i C_{c_2^\prime,c_i^\prime}- \lambda \zeta y_2  r_1^\prime r_1^\prime + \lambda \zeta y_1  r_1^\prime r_2^\prime  \\ &- \lambda y_1 y_2 \sum\limits_{i=3}^{m}  r_1^\prime r_i^\prime  + \lambda \sum\limits_{i=3}^{m}  r_2^\prime  r_i^\prime  - y_1 y_2 +1.
		\end{aligned}
	\end{equation} 
According to equations (\ref{con4})-(\ref{con5}), ${\alpha^{new}_1}^ \prime$ and ${\alpha^{new}_2}^ \prime$ that satisfy the boundary conditions are obtained. In addition, referring to Eq. (\ref{lam}), the parameter $(\lambda^\prime)^t$ of the nonlinear approximate function solution can be updated as follows:
		\begin{equation}\label{lam_1}
		\begin{aligned}
	(\lambda^\prime)^t =\frac{ 2\left\| \sum\limits_{i=1}^{m}\alpha^{t-1}_i y_i c_i^\prime \right\| - \sum\limits_{i=1}^{m}\alpha^{t-1}_i r_i^\prime}{\sum\limits_{i=1}^{m}\alpha^{t-1}_i r_i^\prime} 
		\end{aligned}
\end{equation}}
  In summary, the kernel function can be selected according to different situations to obtain the classification results of the nonlinear kernel function. The detailed solution process is shown in Algorithm \ref{Algorithm1}.
  
   \textcolor{black}{\textbf{Remark:} In step 1 of Algorithm \ref{Algorithm1}, removing overlapping granular-balls with different labels (called heterogeneous granular-balls) is a critical step, especially for the nonlinear GNSVM method. Since the nonlinear model maps data points into a higher dimensional space, the data becomes sparser than before. So heterogeneous granular-balls are likely to occur, which is not conducive to classification. Therefore, it should be noted that removing the overlap of heterogeneous granular-balls in the kernel space can describe the decision boundary more clearly. In the file ``Supplementary\_Material.pdf'', taking the data set balance-scale as an example, the change in the number of granular-balls and accuracy before and after removing overlaps is shown. The results show that removing overlaps produces more granular-balls and greatly improves classification accuracy.}
   

	\begin{algorithm}[tb]\label{Algorithm1}
		\caption{GBSVM optimized using SMO}
		\label{Alg:algoritm2}
		\hspace*{0.02in} {\bf Input:} 
		$\textcolor{black}{D=\{ (x_i,y_i), i=1,2,\dots, n, y_i \in \{+1,-1\} \}}$, the purity threshold $T$ \\
		\hspace*{0.02in} {\bf Output:}
		$\omega$, $b$
		\begin{algorithmic}[1]
			\STATE Generate a granular-ball set ${GB}^{\prime}=\left\lbrace GB_1,GB_2,\dots,GB_m\right\rbrace$ on $D$, where $m$ denotes the number of granular-balls in ${GB}^{\prime}$. 
			\STATE Take the initial vector $\boldsymbol{\alpha}=\textbf{0}$. The optimization objective function is $\frac{1}{2} (\|A\|^2 - \lambda  B^2 )  - \sum\limits_{i=1}^{m}\alpha_i $, where $A = \sum\limits_{i=1}^{m}\alpha_i y_i c_i$ and $B = \sum\limits_{i=1}^{m}\alpha_i r_i$;
			\STATE  \textcolor{black}{Select optimization variables $\alpha_1,\alpha_2$, and solve the two variables according to Eqs. (\ref{con4}) and (\ref{con5}). Parameter $\lambda$ is updated by Eq. (\ref{lam}) for the linear GBSVM model, while the update of $\lambda$ refers to Eq. (\ref{lam_1}) for the nonlinear GBSVM model;}
			\STATE Adjust all the $\alpha_i$ to fulfill the constraints;   
			\IF {The value of the optimized function based on the tuning parameter is reduced}
			\STATE Go to STEP 4 \textbf{else}
			\STATE The optimization is terminated, and go to STEP 9
			\ENDIF
			\STATE According to Eq. (\ref{equ:w'}), calculate $\omega$;
			\STATE According to Eq. (\ref{equ:l'}), calculate $b$.
		\end{algorithmic}
	\end{algorithm}

	\section{Experiment}{\label{sec:experiment}}
	
\begin{table}[!ht]
	\centering
	\setlength{\tabcolsep}{9.5mm}{
		\caption{Dataset Information}
		\label{tab:1}
		\begin{tabular}{lcc}
			\hline
			dataset 	& samples 	& class \\ \hline
			fourclass 	& 862 		& 2 \\ 
			titanic 	& 2201 		& 2 \\ 
			monks-2 	& 432 		& 6 \\ 
			heart1 		& 294 		& 13 \\ 
			haberman 	& 306 		& 3 \\ 
			balance-scale & 625 	& 3 \\ 
			cleveland 	& 303 		& 12 \\ 
			phoneme 	& 5404 		& 5 \\ \hline
	\end{tabular}}
\end{table}

\begin{table*}[htbp!]
	\centering
	\caption{Comparison of accuracy between different methods with different levels of class noise}
	\label{tab:robust}
	\resizebox{\linewidth}{!}{
		\begin{tabular}{llllllllllllllllllllllll}
			\hline
			\multicolumn{2}{l}{Dataset}    & Fourclass & titanic & monks-2 & heart1 & haberman & balance-scale & cleveland & phoneme \\ \hline
			\multirow{3}{*}{0\%}  
			& SVM\cite{platt1998sequential}    & 0.7108    & \textbf{0.7740}   & \textbf{0.6315} & \textbf{0.7734}  & \textbf{0.7623}   & 0.9200 & \textbf{0.8238} & 0.7657 \\
			& GBSVM(PSO\cite{flake2002efficient})   & 0.7139    & ---   & 0.4853 &  0.7458  & 0.7137       & \textbf{0.9540}  & 0.8083 & ---  \\
			& GBSVM(SMO) & \textbf{0.7659}    &0.7466   & 0.5809 & 0.7669  & 0.7500  &  0.9420  & 0.7708 & \textbf{0.7824} \\
			\rule{0pt}{10pt}\multirow{3}{*}{5\%}  
			& SVM\cite{platt1998sequential}    & 0.6909    & 0.7499   & \textbf{0.6352} & \textbf{0.7681}   & 0.7408  & 0.8820 & 0.7880 & 0.7427 \\
			& GBSVM(PSO\cite{flake2002efficient})   & 0.7442    & ---    & 0.4926 & 0.7542  & 0.6492   &  0.9380  & \textbf{0.7917} & --- \\
			& GBSVM(SMO) & \textbf{0.7803}    & \textbf{0.7545}   & 0.6103 & 0.7500 & \textbf{0.7984}  & \textbf{0.9440}  & \textbf{0.7917} & \textbf{0.7572} \\    		
			\rule{0pt}{10pt} \multirow{3}{*}{10\%} 
			& SVM\cite{platt1998sequential}    & 0.6865    & 0.7170   & \textbf{0.6130} & 0.7415  & 0.7029   & 0.8405 & 0.7605 & 0.7140  \\
			& GBSVM(PSO\cite{flake2002efficient})   & 0.7312    & ---    & 0.5147 & 0.7415  & 0.7177   & 0.9500 & 0.7417 & ---  \\
			& GBSVM(SMO) & \textbf{0.7558}    & \textbf{0.7749}   & 0.5515 &  \textbf{0.7585}  & \textbf{0.8065}      & \textbf{0.9620}  & \textbf{0.8250} & \textbf{0.7738}  \\    			
			\rule{0pt}{10pt} \multirow{3}{*}{15\%} 
			& SVM\cite{platt1998sequential}    & 0.6698    & 0.6959   & 0.6019 & 0.7000  & 0.6895    & 0.8025 & 0.7194 & 0.6900 \\
			& GBSVM(PSO\cite{flake2002efficient})   & 0.7327    & ---    & 0.5294 & \textbf{0.7415}  & 0.6895   & 0.9180  & 0.7292 & ---  \\
			& GBSVM(SMO) & \textbf{0.7890}    & \textbf{0.7778}  & \textbf{0.6471}  & 0.7373 & \textbf{0.7661}  & \textbf{0.9340}   & \textbf{0.8375} & \textbf{0.7687} \\
			\rule{0pt}{10pt} \multirow{3}{*}{20\%} 
			& SVM\cite{platt1998sequential}    & 0.6578    & 0.6687   & 0.6019 & 0.6851  & 0.6568  & 0.7475 & 0.6962 & 0.6520 \\
			& GBSVM(PSO\cite{flake2002efficient})   & 0.7500    & ---    & 0.5221 & \textbf{0.7669}  & 0.6694   & 0.9080  & 0.6958 & ---  \\
			& GBSVM(SMO) & \textbf{0.7630}    & \textbf{0.7749}   & \textbf{0.6250} &  0.7542  & \textbf{0.7581}  & \textbf{0.9520}  & \textbf{0.8292} & \textbf{0.7706}\\
			\rule{0pt}{10pt} \multirow{3}{*}{25\%} 
			& SVM\cite{platt1998sequential}    & 0.6306    & 0.6389   & 0.5963 & 0.6404  & 0.6230   & 0.7110     & 0.6709 & 0.6373 \\
			& GBSVM(PSO\cite{flake2002efficient})   & 0.7355    & ---    & 0.4926 & 0.6568  & 0.5887   & 0.9400 & \textbf{0.8083} & --- \\
			& GBSVM(SMO) & \textbf{0.7327}    & \textbf{0.7800}  & \textbf{0.6250} & \textbf{0.7542}  & \textbf{0.7137}   & \textbf{0.9540} & 0.7833 & \textbf{0.7669}  \\
			\rule{0pt}{10pt} \multirow{3}{*}{30\%} 
			& SVM\cite{platt1998sequential}    & 0.6092    & 0.6180   & 0.5852 & 0.6277  & 0.5994   & 0.6610 & 0.6392 & 0.6070 \\
			& GBSVM(PSO\cite{flake2002efficient})   & 0.7153    & ---    & 0.5662 & \textbf{0.7161}  & 0.6492   & \textbf{0.9220} & 0.5833 & --- \\
			& GBSVM(SMO) & \textbf{0.7384}  & \textbf{0.7710}  & \textbf{0.6471} & 0.7034 & \textbf{0.7218}  & 0.9040 & \textbf{0.7833}  & \textbf{0.7623} \\ \hline
		\end{tabular}
	}
\end{table*}

	Since most data sets are not separable in practice, in this section, the GBSVM is compared with different models in the performance of robustness and efficiency. The relative density (RD) method \cite{xia2021mcrf} is introduced to compare the classification results of data sets with different noise levels. The method is designed with relative density based on absolute density by calculating the ratio of the distances between the k-nearest heterogeneous neighbors and k-nearest homogeneous neighbors of each sample to eliminate noisy data. In addition, if RD completely filters out a class of data points, we cancel the noise filtering step and take the original data. Then, SVM \cite{platt1998sequential} and GBSVM utilizing SMO algorithm are compared with GBSVM employing PSO algorithm \cite{flake2002efficient}. Furthermore, we added a comparison experiment between SVM and GBSVM with the RD framework. For fair comparisons, all parameters are set to the same. Several UCI benchmark datasets are randomly selected in our experiments. The dataset information is shown in Table \ref{tab:1}. Experimental hardware environment is shown as follows: PC with an Intel Core i7-107000 CPU @2.90 GHz with 32 G RAM. Experimental software environment: Python 3.7.

 \begin{table*}[htbp!]
 	\centering
 	\caption{\textcolor{black}{The results of GBSVM with Gaussian kernel on different datasets.}}
 	\label{tab:gauss}
 	\resizebox{\linewidth}{!}{
 		\begin{tabular}{llllllllllllllllllllllll}
 			\hline
 			\multicolumn{1}{l}{Dataset}  &  & Fourclass & titanic & monks-2 & heart1 & haberman & balance-scale & cleveland & phoneme \\ \hline
 			\rule{0pt}{10pt}\multirow{2}{*}{ 0\%}  
 			& SVM\cite{platt1998sequential}  & \textbf{0.7211}    & 0.7664     & 0.5735  & 0.7415   & 0.7500 & 0.9440 & 0.8042 & 0.7606 \\
 			& GBSVM(SMO)  & 0.7081    & \textbf{0.7681}   & \textbf{0.6176}  & \textbf{0.7458}  & \textbf{0.7540}   & \textbf{0.9520} & \textbf{0.8333} & \textbf{0.7631} \\
 			\rule{0pt}{10pt}\multirow{2}{*}{   5\%}  
 			& SVM\cite{platt1998sequential}  & \textbf{0.7355}    & 0.7755    &0.5735  & \textbf{0.7542} & 0.7379  & 0.9360 & 0.7667  & 0.7120 \\
 			& GBSVM(SMO) & 0.6705  & \textbf{0.7972}    & \textbf{0.6176} & 0.7458   & \textbf{0.7581}     & \textbf{0.9520}  & \textbf{0.8333} & \textbf{0.7418}\\
 			\rule{0pt}{10pt} \multirow{2}{*}{10\%}
 			& SVM\cite{platt1998sequential}  & \textbf{0.7456}     & 0.7021   & 0.5809  & 0.7373   & 0.7258  & 0.9360 & 0.7875 & 0.7168 \\
 			& GBSVM(SMO) & 0.7399   & \textbf{0.7578}   & \textbf{0.6397}  & \textbf{0.7797}  & \textbf{0.7419}   & \textbf{0.9680}  & \textbf{0.8167} & \textbf{0.7448}  \\
 			\rule{0pt}{10pt} \multirow{2}{*}{15\%} 
 			& SVM\cite{platt1998sequential}  & 0.7201    & 0.6927    & 0.6471  & 0.7246   & 0.7823 & \textbf{0.9400} & 0.7500 & 0.6794 \\
 			& GBSVM(SMO) & \textbf{0.7283}    & \textbf{0.7914}   & \textbf{0.5882}  & \textbf{0.7627}  & \textbf{0.7903}   & 0.9280   & \textbf{0.8167} & \textbf{0.7572} \\
 			\rule{0pt}{10pt} \multirow{2}{*}{20\%} 
 			& SVM\cite{platt1998sequential}  & \textbf{0.7919}    & 0.6626    & 0.5662  & 0.7161   & 0.6855  & 0.9480 & 0.7250 & 0.6584 \\
 			& GBSVM(SMO) & 0.7630   & \textbf{0.7732}    & \textbf{0.5809}  & \textbf{0.7797}  & \textbf{0.7419}       & 	\textbf{0.9520}  & \textbf{0.8333} & \textbf{0.7769} \\
 			\rule{0pt}{10pt} \multirow{2}{*}{25\%} 
 			& SVM\cite{platt1998sequential}  & \textbf{0.7933}    & 0.6411    & 0.5441  & \textbf{0.7839}   & 0.7500  & 0.9040 & 0.6958 & 0.6080 \\
 			& GBSVM(SMO)  & 0.7168    & \textbf{0.7937}    & \textbf{0.6765}  & 0.7627  & \textbf{0.7903}       & \textbf{0.9520}  & \textbf{0.7500} & \textbf{0.7980} \\
 			\rule{0pt}{10pt} \multirow{1}{*}{30\%} 
 			& SVM\cite{platt1998sequential}  & 0.7630   & 0.6161    & \textbf{0.5956}  & 0.7246   & 0.6895  & 0.9040 & 0.6625 & 0.6145 \\
 			& GBSVM(SMO)  & \textbf{0.7803}    & \textbf{0.7891}    & \textbf{0.5956}   & \textbf{0.7542}  & \textbf{0.7379}       & \textbf{0.9520}  & \textbf{0.7333} & \textbf{0.7130} \\ \hline
 		\end{tabular}
 	}
 \end{table*} 
 
	The parameters about PSO are set as follows: the dimension $dim$ is equal to the number of the conditional attributes; the number $pop$ of particles is 400; the maximum number of iterations $max \_iter$ is 1050; inertial factor $w$ is set to 0.5; learning factors both $c_1$ and $c_2$ equal to 1.6. For each dataset, we conducted training and testing in an 8:2 ratio. In each dataset experiment, we choose the optimal model based on the training set evaluation and then input the test set to obtain results. The experimental results of GBSVM and SVM are the average results of running four times on each dataset. To ensure a fair comparison, the parameter $C$ is set to $1.0$. The results of solving the model using SMO and PSO algorithms are compared, and the experimental results using the nonlinear kernel are provided.
 
 			\begin{figure}[htbp!]
 			\centering
 			\subfigure[noise=0, purity threshold=1]{\includegraphics[width=0.24\textwidth,height=0.2\textwidth]{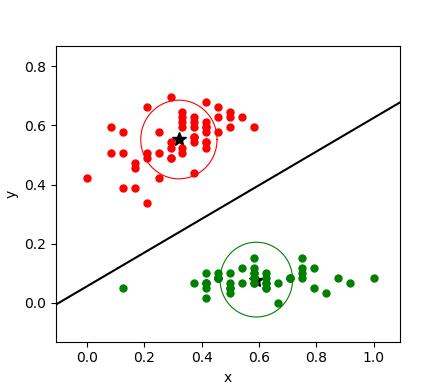}}
 			\subfigure[noise=5\%, purity threshold=0.98]{\includegraphics[width=0.24\textwidth,height=0.2\textwidth]{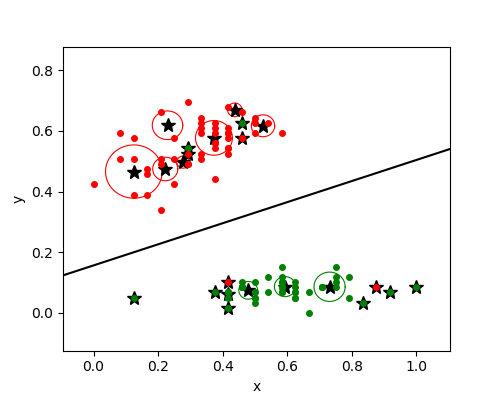}}
 			\subfigure[noise=0, purity threshold=1]{\includegraphics[width=0.24\textwidth,height=0.2\textwidth]{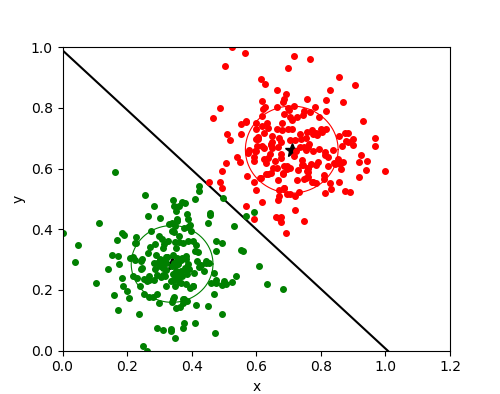}}
 			\subfigure[noise=5\%, purity threshold=0.98]{\includegraphics[width=0.24\textwidth,height=0.2\textwidth]{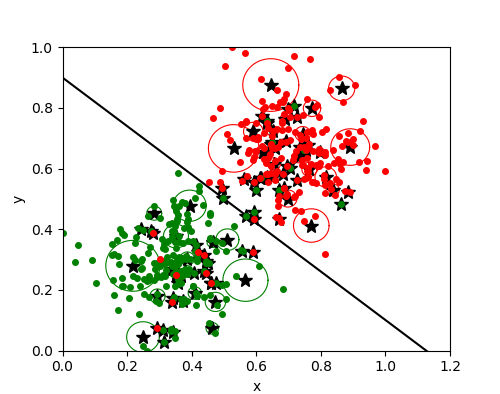}}
 			\caption{The results of linear GBSVM for Iris data set and a newly generated dataset.}
 			\label{fig:6-7}
 		\end{figure}

 		First, we conducted visualization experiments using both the classic dataset Iris and a newly generated dataset, as depicted in Fig. \ref{fig:6-7}. Here, the black straight line represents the decision plane. The first two dimensions of the dataset Iris are chosen to analyze the experimental results before and after the introduction of noise. In Fig. \ref{fig:6-7}(a) and (b), the noise level is set at 0\% and 5\%, and the purity thresholds are set at 1 and 0.98. Results show that GBSVM model derives a superior decision plane. Due to the limited size of the dataset Iris, a newly generated dataset was used for further testing. The results confirmed that small noise will not affect the granular-ball describing the data distribution, ensuring that the GBSVM model can distinguish the two class data.

	In the experiments, each dataset is contaminated with different percentages of label noise, including 0$\%$, 5$\%$, 10$\%$, 15$\%$, 20$\%$, 25$\%$ and 30$\%$. It is important to note that we only introduced noise to the training set while maintaining the test set free of noise. The accuracy comparison is shown in Table \ref{tab:robust}. The more accurate algorithm on each dataset is marked in \textbf{boldface}. With the increasing level of label noise, SVM shows good performance in robustness on most datasets. However, GBSVM shows better robustness than SVM. The bold numbers in table \ref{tab:robust} mostly appear in the results of GBSVM solved by PSO or SMO algorithm. It indicates that GBSVM can achieve a higher accuracy than SVM at the same level of label noise. The reason is that the label noise points do not affect the label of a granular-ball, which is determined by the majority label in the granular-ball. Since the "titanic" and "phoneme" data sets are large and use the PSO algorithm requires a long optimization time, the experimental results cannot be obtained within one day. Consequently, we did not document the final outcomes. Furthermore, we visualized the results in the file ``Supplementary\_material.pdf''.
	In fact, SMO algorithm can get higher classification accuracy than PSO method. As noise increases, some results exhibit improved precision as the noise level rises. Some results improves with increasing noise levels, which can be attributed to the absence of noise in the test set or the final result being the average of four trials. Moreover, we also provide the GBSVM results using ``Gaussian'' kernel in Table \ref{tab:gauss} and compare them with the SVM solved by SMO algorithm. \textcolor{black}{The results indicate that non-linear GBSVM still has higher classification accuracy than SVM on most datasets. However, as the overlap of heterogeneous granular-balls in kernel space is removed, the number of granular-balls increases. Taking the dataset balance-scale as an example, the table in the file ``Supplementary-Material. pdf'' shows the changes in both the number of granular-balls and the accuracy, before and after removing overlaps at varying purities. The results show that although removing overlaps increases the number of granular-balls, it greatly improves the classification accuracy.} In short, the accuracy advantage of GBSVM is more obvious at high noise levels.
	
	Since the experiments in this work mainly focus on label noise, the performance of GBSVM is further demonstrated with some denoising methods, such as the RD method, as shown in the file ``Supplementary\_Material.pdf''. The results show that RDGBSVM achieves better results on most datasets. Actually, the GBSVM model with granular-balls instead of points as input is not designed specifically for noisy data, but has its own advantages of efficiency and robustness. This is a self-contained advantage of the model using granular-balls computing, which is also scalable.
	\begin{figure}[!ht]
		\setlength{\abovecaptionskip}{-0.1cm}  
		\centering
		\subfigure[]{\includegraphics[width =1.6in]{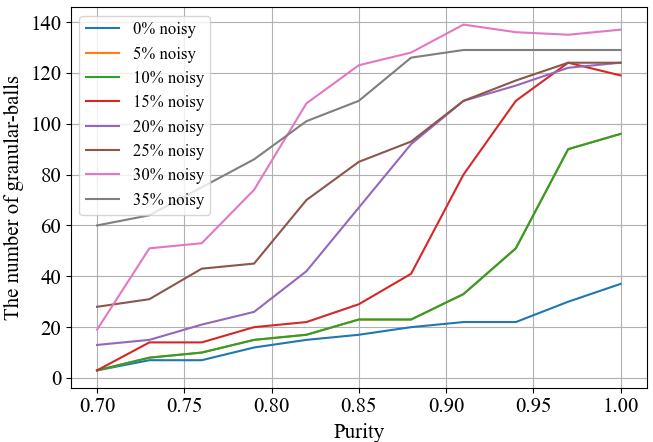}}
		\subfigure[]{\includegraphics[width =1.6in]{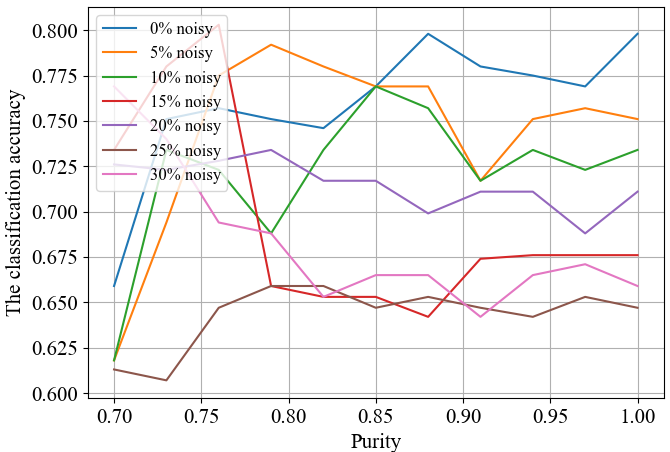}}
		\caption{Take the dataset Fourclass as an example, the number of granular-balls and the classification accuracy are generated by using linear GBSVM method with different purities. (a) The number of granular-balls; (b) The classification accuracy.} 
		\label{fig:GB}
	\end{figure}
	
	Taking the dataset Fourclass as an example, Fig \ref{fig:GB} shows the number of granular-balls and the classification accuracy using linear GBSVM method with different purity thresholds, and SMO is used as the solution algorithm. In order to improve the computational efficiency, we eliminate the granular-balls containing only one isolated point. It can be clearly seen that the smaller the purity threshold, the smaller the number of granular-balls generated in Fig. \ref{fig:GB}(a). As shown in Fig. \ref{fig:GB}(b), appropriately reducing the purity threshold can improve the classification accuracy in the case of high noise. The higher the noise, the lower the optimal threshold. The reason is that a smaller purity threshold allows the granular-ball to contain more noise points without affecting the label of the entire ball, making GBSVM more robust.

	For a fair comparison of efficiency, both SVM and GBSVM are solved by the SMO algorithm under the same setting of parameters, and compared with GBSVM using PSO algorithm. The experimental results are shown in Table \ref{tab:efficiency}. The fastest algorithm on each dataset is marked in \textbf{boldface}. In this experiment, the purity threshold is optimized in the interval [0.7,1] with a step size of 0.015, and the running time in the table corresponds to the running time of the optimal result. The reference \cite{2021GBAdai} can adaptively generate granular-balls without optimizing the purity threshold, and the time required is very short. The experimental results show that GBSVM is dozens or hundreds of times faster than SVM. The reason is that the number of granular-balls is much smaller than that of the original samples, making GBSVM much faster than SVM. Moreover, each iteration of the Lagrange multiplier corresponds to a search process for PSO method; in contrast, the SMO algorithm is much more efficient than the PSO algorithm by solving the analytical solution. In conclusion, GBSVM is a more efficient and robust model than SVM model.
 		\vspace{-0.4cm}
	\begin{table}[!ht]
		\centering
		\caption{Comparison of the running time between GBSVM and SVM}
		\label{tab:efficiency}
		\setlength{\tabcolsep}{2.8mm}{
			\begin{tabular}{lrrr}
				\hline
				dataset    & \multicolumn{1}{c}{SVM\cite{platt1998sequential}}   &  \multicolumn{1}{c}{GBSVM(PSO\cite{flake2002efficient})}		   &  \multicolumn{1}{c}{GBSVM(SMO)}   \\ \hline
				fourclass     & 0.2231       & 96845.63    & \textbf{0.0800}  \\ 
				titanic       & 0.4800     & 277498.88    & \textbf{0.0082}        \\
				monks-2       & \textbf{0.1341}      & 59455.78   &  9.1882      \\
				heart1        & 0.0786      & 17959.23  & \textbf{0.0020}  \\
				haberman      & 0.0896      & 34687.36  & \textbf{0.0598}  \\
				balance-scale & 0.1131      & 89012.17  & \textbf{0.0758}   \\
				cleveland     & 0.1163     & 16855.56  & \textbf{0.0059}     \\
				phoneme       & 4.7237	 & 746588.43 & \textbf{3.1118}    \\ \hline
		\end{tabular}}
	\end{table}

	\section{Conclusion and Future Work} {\label{sec:conclusion}}

		The proposed GBSVM is an important attempt to use the different granularities of granular-balls as the input to construct a classifier. GBSVM is the first nonpoint input classifier in the history of machine learning, and its model does not contain points, that is, $x_i$. This paper has fixed the errors in the existing GBSVM, and derived its dual model. The nonlinear GBSVM model is proposed by developing a method of generating granular-balls in kernel space. An approximate function solving algorithm based on the SMO algorithm is designed to calculate the dual model. The experimental results on some benchmark datasets demonstrate that GBSVM performs better than SVM both in robustness and effectiveness. It also indicates that granular-ball classifiers perform well in robustness and effectiveness.

		Although GBSVM exhibits good performance in efficiency and robustness, many drawbacks need to be overcome such as how to accurately find the solution of the model. This paper only presents a basic framework, and the classification based on granular-ball computation can be extended to more existing advanced models, which are widely used in the field of pattern recognition, such as face recognition, handwritten character recognition, speech recognition, image recognition, etc. In fact, there are also requirements for high efficiency and robustness in automation and other fields, such as \cite{guan2023spline,ning2023practical,xiao2023sampled,bing2023towards,huang2023cost,guan2023autonomous,xuan2023practical}. So, developing granular-ball learning into other computing methods \cite{hossain2023msfired,xiao2023sampled,hu2022review} will be an important work in the future.
	
	
	
	\ifCLASSOPTIONcaptionsoff
	\newpage
	\fi
	
	\bibliographystyle{unsrt}
	\bibliography{references}

\end{document}